\documentclass[review]{elsarticle}

\usepackage{hyperref}
\usepackage{longtable}
\usepackage{amsmath,amsthm,amssymb,epsfig,bm,amsmath}
 \usepackage{graphicx}
\usepackage{float}
\usepackage{amsmath}
\usepackage{mhchem}
\usepackage{xparse}
\usepackage{adjustbox}
\usepackage{MnSymbol}
\usepackage{mathdots}
\usepackage{makecell}
\usepackage{listings}
\usepackage{footnote}
\usepackage[flushleft]{threeparttable}
\usepackage{multirow}
\usepackage{relsize}
\usepackage{lettrine}
\usepackage{courier}
\usepackage{color} 
\definecolor{mygreen}{RGB}{28,172,0} 
\definecolor{mylilas}{RGB}{170,55,241}
\usepackage{amsmath,amsthm,amssymb,amsfonts} 
\usepackage{graphicx} 
\usepackage[margin=1in]{geometry} 
\usepackage[yyyymmdd]{datetime}
\usepackage{algorithm}
\usepackage[noend]{algpseudocode}
\usepackage{lipsum}
\usepackage{setspace}
\usepackage{siunitx}
\usepackage{pdfpages}
\usepackage{tikz}
\usetikzlibrary{positioning}
\usepackage{booktabs,caption}
\usepackage[]{hyperref}
\usepackage{soul}
\usepackage{xcolor}
\usepackage{subcaption}

\hypersetup{
 colorlinks  = true, 
 urlcolor   = blue, 
 linkcolor  = black, 
 citecolor  = blue 
}
\usepackage[numbers]{natbib}
\usepackage{color}
\definecolor{codegreen}{rgb}{0,0.6,0}
\definecolor{codegray}{rgb}{0.5,0.5,0.5}
\definecolor{codepurple}{rgb}{0.58,0,0.82}
\definecolor{backcolour}{rgb}{0.95,0.95,0.92}
\lstdefinestyle{mystyle}{
  backgroundcolor=\color{backcolour},  
  commentstyle=\color{codegreen},
  keywordstyle=\color{magenta},
  numberstyle=\tiny\color{codegray},
  stringstyle=\color{codepurple},
  basicstyle=\footnotesize,
  breakatwhitespace=false,     
  breaklines=true,         
  captionpos=b,          
  keepspaces=true,         
  numbers=left,          
  numbersep=5pt,         
  showspaces=false,        
  showstringspaces=false,
  showtabs=false,         
  tabsize=2,
  escapeinside={<@}{@>},
}
 
\usepackage{booktabs} 
\usepackage{siunitx} 
\usepackage{comment}
\usepackage{mathtools} 
\usepackage{gensymb} 
\usepackage{mhchem} 
\usepackage{fixltx2e} 
\usepackage{bbm}

\usepackage{multirow}

\usepackage{fancyhdr} 
\theoremstyle{definition}

\theoremstyle{definition}

\theoremstyle{remark}

\usepackage{soul} 
\soulregister\cite7
\soulregister\ref7
\soulregister\pageref7
\usepackage{bm}

\usepackage{framed} 

\usepackage{multicol} 

\usepackage{nomencl} 
\usepackage{tikz}
\makenomenclature
\usepackage[resetlabels,labeled]{multibib}

\renewcommand*\nompreamble{\begin{multicols}{2}}
\renewcommand*\nompostamble{\end{multicols}}

\usepackage{amssymb}
\usepackage{pifont}
\definecolor{light-gray}{gray}{0.95}

\makeatother

\usepackage[left, pagewise]{lineno}

\journal{Elsevier}

\begin{document}


\begin{frontmatter}

\title{\large Multifidelity Surrogate Modeling of Depressurized Loss of Forced Cooling in High-temperature Gas Reactors}

\author{Meredith Eaheart$^{a*}$, Majdi I. Radaideh$^{a,b*}$}

\cortext[mycorrespondingauthor]{Corresponding Author: M. Eaheart (eaheart@umich.edu), M. I. Radaideh (radaideh@umich.edu)}

\address{$^{a}$Department of Nuclear Engineering and Radiological Science, University of Michigan, Ann Arbor, Michigan 48109}
\address{$^{b}$Department of Computer Science and Engineering, University of Michigan, Ann Arbor, Michigan 48109}

\begin{abstract}
\small
High-fidelity computational fluid dynamics (CFD) simulations are widely used to analyze nuclear reactor transients, but are computationally expensive when exploring large parameter spaces. Multifidelity surrogate models offer an approach to reduce cost by combining information from simulations of varying resolution. In this work, several multifidelity machine learning methods were evaluated for predicting the time to onset of natural circulation (ONC) and the temperature after ONC for a high-temperature gas reactor (HTGR) depressurized loss of forced cooling transient. A CFD model was developed in Ansys Fluent to generate 1000 simulation samples at each fidelity level, with low and medium-fidelity datasets produced by systematically coarsening the high-fidelity mesh. Multiple surrogate approaches were investigated, including multifidelity Gaussian processes and several neural network architectures, and validated on analytical benchmark functions before application to the ONC dataset. The results show that performance depends strongly on the informativeness of the input variables and the relationship between fidelity levels. Models trained using dominant inputs identified through prior sensitivity analysis consistently outperformed models trained on the full input set. The low- and high-fidelity pairing produced stronger performance than configurations involving medium-fidelity data, and two-fidelity configurations generally matched or exceeded three-fidelity counterparts at equivalent computational cost. Among the methods evaluated, multifidelity GP provided the most robust performance across input configurations, achieving excellent metrics for both \textit{time to ONC} and {temperature after ONC}, while neural network approaches achieved comparable accuracy with substantially lower training times. These results demonstrate that multifidelity surrogate models can reduce the computational burden of reactor transient analysis while maintaining accurate prediction of key safety metrics.

\end{abstract}

\begin{keyword}
Multifidelity Modeling, Computational Fluid Dynamics (CFD), Nuclear Reactor Safety, High Temperature Gas Reactors, Applied Machine Learning
\end{keyword}

\end{frontmatter}


\setstretch{1.3}


\section{Introduction}
\label{sec:intro}
High-temperature gas-cooled reactors (HTGRs) can experience transients such as depressurized loss of forced cooling (DLOFC) accidents in which a break in the primary pressure boundary allows air to ingress into the helium-filled system \cite{oh2011airingress}. This process is not instantaneous: as air accumulates and a critical concentration is reached, natural circulation can commence and bring air into the system at a much higher rate \cite{gould2017transition,takeda1992studies}. Once natural circulation develops, the increased air ingress can oxidize graphite structures and degrade core heat removal capability, so the time to onset of natural circulation (ONC) is an important safety metric for HTGR analysis \cite{franken2018numerical}. 

Accurate prediction of ONC timing in DLOFC scenarios requires high-fidelity computational fluid dynamics (CFD) simulations that resolve fluid behavior over long transients of several thousand seconds of physical flow time~\cite{franken2018numerical}. Such simulations are computationally expensive, limiting the number of runs that can be afforded for tasks such as sensitivity analysis \cite{radaideh2020efficient}, uncertainty quantification \cite{price2019advanced}, and design optimization \cite{radaideh2021neorl,radaideh2023neorl}. For detailed three-dimensional geometries, running hundreds or thousands of full high-fidelity transients is often infeasible in practice, which motivates the use of surrogate and multifidelity machine learning approaches to approximate time to ONC from a limited set of high-fidelity simulations.

Several experiments~\cite{gould2017transition,takeda1992studies} and simulations~\cite{franken2018numerical,eaheart2024effect,yurko2008effect,eaheart2025sensitivity} have been conducted to determine the time to ONC for DLOFC transients in HTGRs. In recent work, the authors developed an automated HTGR DLOFC CFD workflow and performed sensitivity analysis and global and local uncertainty quantification to determine which inputs most significantly impact the time to ONC using 500 high-fidelity simulations with randomized inputs. Eight thermal and thermophysical parameters were perturbed within physically realistic ranges, and the heated-section temperature was found to be the dominant driver of ONC, with minimal impacts from the remaining perturbed input parameters \cite{eaheart2025sensitivity}. Building on this sensitivity analysis, which identifies both dominant and non-dominant (weakly sensitive) inputs, we test whether two-fidelity and three-fidelity approaches perform differently across these input groups. Multifidelity surrogate methods have not been applied to ONC prediction, and it is unclear how input sensitivity structure affects multifidelity surrogate performance in this setting.

Recent advances in machine learning have enabled the development of surrogate models to accelerate high-fidelity simulations and data analysis in nuclear engineering applications. Early efforts demonstrated how data-driven models can be used to emulate computationally expensive nuclear simulations and support uncertainty quantification and system analysis. For example, the use of the Group Method of Data Handling (GMDH) was explored to analyze reactor simulation data and construct surrogate models capable of capturing nonlinear relationships in reactor physics datasets \cite{radaideh2020analyzing}. Further work introduced probabilistic surrogate modeling approaches such as Gaussian processes to approximate advanced multiphysics simulations while quantifying predictive uncertainty \cite{radaideh2020surrogate,nasr2025enhancing}. In parallel, neural-network surrogates have been applied to time-series prediction of accident scenarios and plant transients \cite{antonello2023physics,radaideh2020neural}, as well as to high-dimensional neutronics problems such as boiling water reactor core modeling \cite{oktavian2024integrating,saleem2020application}. The integration of simulation outputs with machine learning has also been explored to combine physics-based models and data-driven approaches for energy system modeling and uncertainty quantification \cite{radaideh2019combining}. Beyond simulation emulation, surrogate models have enabled calibration and optimization tasks in nuclear systems, including evolutionary neural network and polynomial surrogate approaches for spallation target calibration \cite{radaideh2022model} and surrogate-assisted optimization of reactor operation and design problems \cite{sobes2021ai,price2022multiobjective}. Similar developments have been reported in the literature, including machine-learning-based surrogates to accelerate core optimization with fuel performance modeling \cite{che2022machine} and to speed up genetic algorithm optimization for coupled fast/thermal nuclear experiments \cite{pevey2022neural}. More recently, data-driven surrogate frameworks have been extended to operational intelligence applications such as early fault detection in accelerator power electronics \cite{radaideh2023early} and reinforcement-learning-based control and power shaping in nuclear microreactors \cite{radaideh2025multistep}. Collectively, these studies highlight the growing role of surrogate modeling and machine learning in enabling faster simulation, optimization, monitoring, and decision support for advanced nuclear energy systems.

Multifidelity methods combine information from at least two models or data sources that differ in computational cost and expected accuracy, typically referred to as low-fidelity (LF) and high-fidelity (HF). In this work, LF denotes the cheaper and less accurate model or data set and HF denotes the most accurate and more expensive one, with “fidelity” defined relative to available modeling options rather than as absolute truth \cite{fernandez2016review,peherstorfer2018survey}. Different types of low-high-fidelity combinations can be used such as low vs. higher dimensionality ~\cite{turner2004multi,wang2013novel}, coarse vs. fine meshes ~\cite{jonsson2015shape, toal2015some}, simulation vs. experimental results~\cite{kuya2011multifidelity}, and steady-state vs. transient simulations~\cite{berci2011multifidelity}. Multifidelity methods are most effective when there is a strong correlation between LF and HF outputs, such that the LF model captures the dominant trends of the HF response even if it cannot fully resolve all features \cite{fernandez2016review,peherstorfer2018survey}.

Multifidelity methods can be broadly categorized into multifidelity surrogate models (MFSMs) and multifidelity hierarchical models (MFHMs). MFSMs explicitly fuse data from multiple fidelity levels into a single surrogate that directly predicts HF output, such as co-kriging or a multifidelity neural network \cite{fernandez2016review}. MFHMs, in contrast, employ different fidelity levels without constructing an explicit joint surrogate, for example through 
adaptive sampling \cite{choi2008multifidelity}. The present work focuses on MFSMs, which will be the focus of the remainder of this section.

Multifidelity surrogate modeling has been broadly applied in fluid mechanics, including aerodynamic shape optimization \cite{pena2012surrogate,tao2019application}, ship performance prediction\cite{wackers2020multi}, wind load estimation on structures \cite{lamberti2021multi}, and turbomachinery flow field prediction \cite{li2023multi}. It has also been applied in nuclear engineering, including machine-learning-corrected coarse-mesh CFD simulations for thermal-hydraulics applications \cite{hanna2020machine} and reactor core simulations \cite{oktavian2023preliminary}.

Common multifidelity surrogate approaches fall into two broad families: Gaussian process (GP)-based methods and neural network-based methods. GP-based methods use co-kriging to model the relationship between fidelity levels probabilistically \cite{kennedy2000predicting}, and nonlinear extensions have since relaxed the assumption of a linear inter-fidelity mapping \cite{perdikaris2017nonlinear}. Neural network-based approaches learn this mapping through composite or residual architectures \cite{meng2020composite, guo2022multi, yi2024practical}, offering greater flexibility for high-dimensional problems. Multifidelity extensions of DeepONet, a neural-network-based operator learning approach, have also been proposed \cite{demo2023deeponet, yangdata}.

This work develops and evaluates multifidelity machine learning surrogates to predict time to ONC and post-ONC temperature in a transient air ingress of a HTGR DLOFC using three mesh-based fidelity levels derived from the same Ansys Fluent model \cite{eaheart2025sensitivity}. The main contributions of this work are as follows:

\begin{itemize}
    \item Application of multifidelity machine learning surrogates to the prediction of time to ONC and post-ONC temperature in an HTGR DLOFC air-ingress transient, a problem class where such methods have not previously been applied. Such methods have the potential to significantly accelerate the safety analysis paradigm for HTGR DLOFC, which is often hindered by the inherently slow transient behavior and the substantial computational cost required for high-fidelity simulations.
    \item A systematic comparison of two-fidelity and three-fidelity strategies under equivalent computational budgets, evaluating the marginal value of an additional fidelity level across multiple surrogate architectures. Applications are demonstrated on mathematical benchmark functions for method verification and a real-world engineering problem. 
    \item An input partitioning study leveraging prior sensitivity analysis to assess surrogate performance under dominant, non-dominant, and full input parameter sets, demonstrating the impact of input selection on multifidelity surrogate accuracy.
\end{itemize}

The remainder of this paper is organized as follows. Section~\ref{sec:methods} describes the CFD simulation and data generation, the multifidelity surrogate methods, and the experimental design. Section~\ref{sec:results} presents benchmark validation and ONC results. Section~\ref{sec:discussion} discusses the findings across input subsets and fidelity configurations, and Section~\ref{sec:conclusions} summarizes conclusions and directions for future work.

\section{Methods}
\label{sec:methods}

\subsection{CFD Simulation and Data Generation}
\label{sec:data}

\subsubsection{Ansys Fluent Simulation}
\label{sec:fluent}
\begin{figure}[h]
    \centering
    \includegraphics[width=0.6\textwidth]{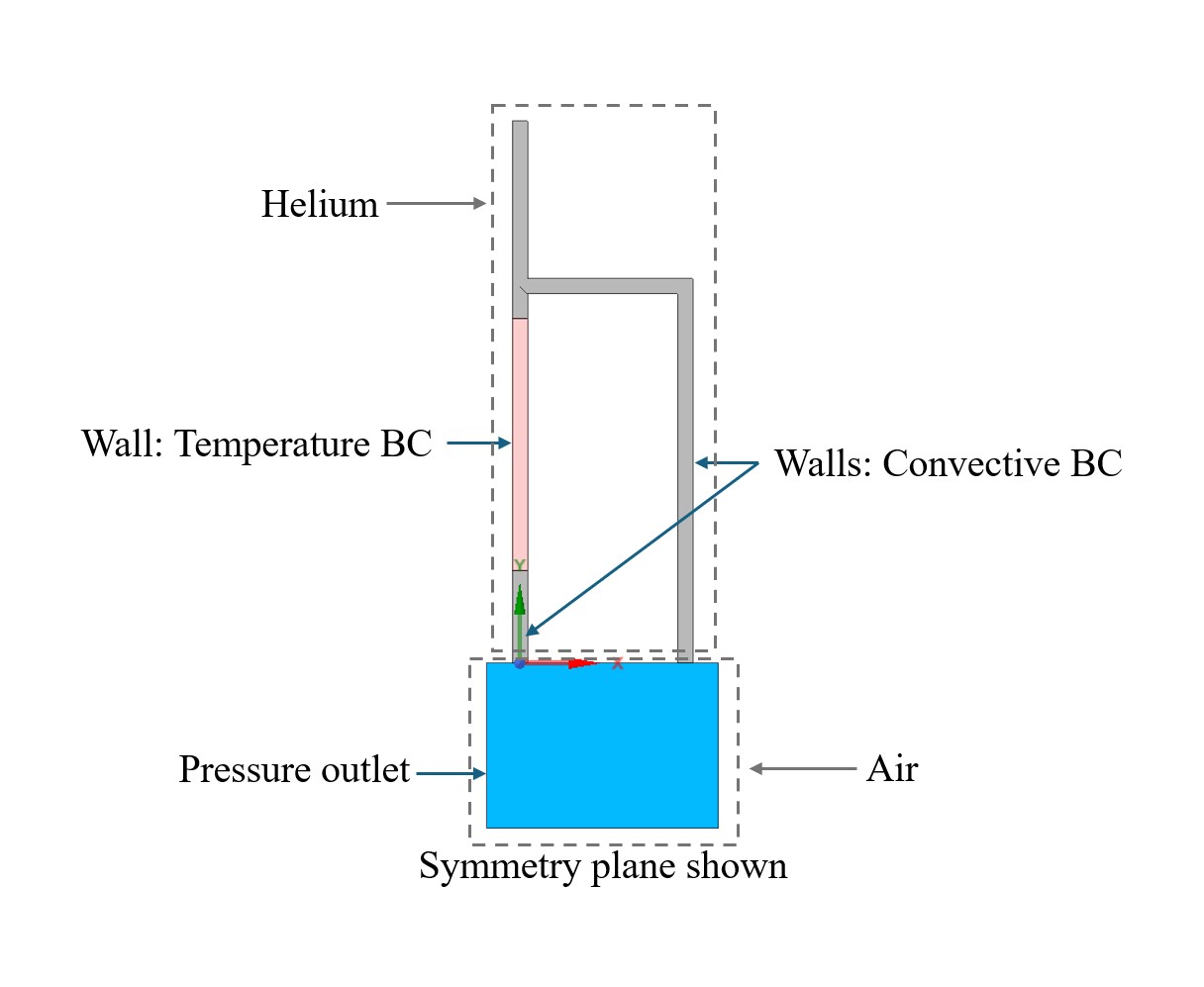}
    \caption{Computational domains with boundary conditions of the Ansys Fluent simulation}
    \label{fig:symmetry_view}
\end{figure}

The computational domain of this simulation represents a simplified h-loop geometry representing an HTGR filled with helium and a bottom section filled with air. The upper h-loop portion has a heated section that employs a constant temperature boundary condition and an unheated section that is modeled with a convective boundary condition with a defined constant heat transfer coefficient as shown in Figure~\ref{fig:symmetry_view}. The lower portion of the domain represents an opening to the atmosphere, modeled using a pressure outlet boundary condition with backflow enabled to allow air-helium mixing.

Initial conditions represent a post-depressurization state, i.e. the pressure is at atmospheric condition at the beginning of the transient. Due to the slow development of air ingress and natural circulation, each transient simulation is run for 6000 seconds of flow time. Transient simulations were performed using Ansys Fluent with a time step of 0.2 seconds. This large time step was chosen as a compromise to ensure stability while enabling the completion of a large amount of simulations given computational resource constraints. Because of this, the time to ONC is meant to serve as a relative value between the different inputs and not meant to give an absolute time to ONC without further model work and validation efforts. Consistent time stepping across all fidelity levels ensures fair comparison of multifidelity learning approaches. Additional information about the simulation can be found in~\cite{eaheart2025sensitivity}.

\subsubsection{Input Parameters and SA/UQ}

Our previous work \cite{eaheart2025sensitivity} performed sensitivity analysis and uncertainty quantification on the Ansys Fluent CFD model using 500 high-fidelity simulations with randomized input parameters. Eight parameters were perturbed across physically realistic ranges to characterize their influence on ONC timing and temperature after ONC including the heated section temperature, the heat transfer coefficient of the unheated section, glass thickness, glass thermal conductivity, air viscosity, air thermal conductivity, helium viscosity, and helium thermal conductivity. These ranges are shown in Table~\ref{tab:param_ranges}.

\begin{table}[h!]
\centering
\caption{Parameter bounds used in the sensitivity and uncertainty analysis. All parameters were sampled using a uniform distribution ~\cite{eaheart2025sensitivity}.}
\label{tab:param_ranges}
\begin{tabular}{lcc}
\hline
\textbf{Parameter} & \textbf{Lower Bound} & \textbf{Upper Bound} \\
\hline
Initial temperature of heated section (K) & 873.15 & 1498.2 \\
Heat transfer coefficient of unheated section (W/m\textsuperscript{2}K) & 0.1 & 10 \\
Air viscosity (kg/m·s)~\cite{keenan1980gastables,touloukian1970tprc} & 1.85$\times$10\textsuperscript{-5} & 5.16$\times$10\textsuperscript{-5} \\
Air thermal conductivity (W/m·K)~\cite{keenan1980gastables,touloukian1970tprc} & 0.02551 & 0.08452 \\
Helium viscosity (kg/m·s)~\cite{nist_helium} & 1.98$\times$10\textsuperscript{-5} & 6.15$\times$10\textsuperscript{-5} \\
Helium thermal conductivity (W/m·K)~\cite{nist_helium}& 0.15525 & 0.47859 \\
Glass thermal conductivity (W/m·K)~\cite{sugwara1969precise} & 1.4 & 3.2 \\
Glass thickness (m) & 0.001 & 0.004 \\
\hline
\end{tabular}
\end{table}

Sensitivity analysis was conducted using four different methods, including Sobol analysis~\cite{sobol2001global}, Fourier amplitude sensitivity testing (FAST)~\cite{mcrae1982global}, morris screening~\cite{Morris01051991}, and regional sensitivity analysis (RSA)~\cite{hornberger1981approach} with both random forest and neural network surrogate models. This was done for both the time to ONC and temperature after ONC to show the influence of the different input parameters on these outputs. For time to ONC, the temperature of the heated section was shown to have the greatest impact with the other input parameters having minimal influence. Both temperature and heat transfer coefficient were shown to have an influence on the temperature after ONC, with minimal impact from the other inputs.

\begin{figure}[h]
\centering
\includegraphics[width=0.7\textwidth]{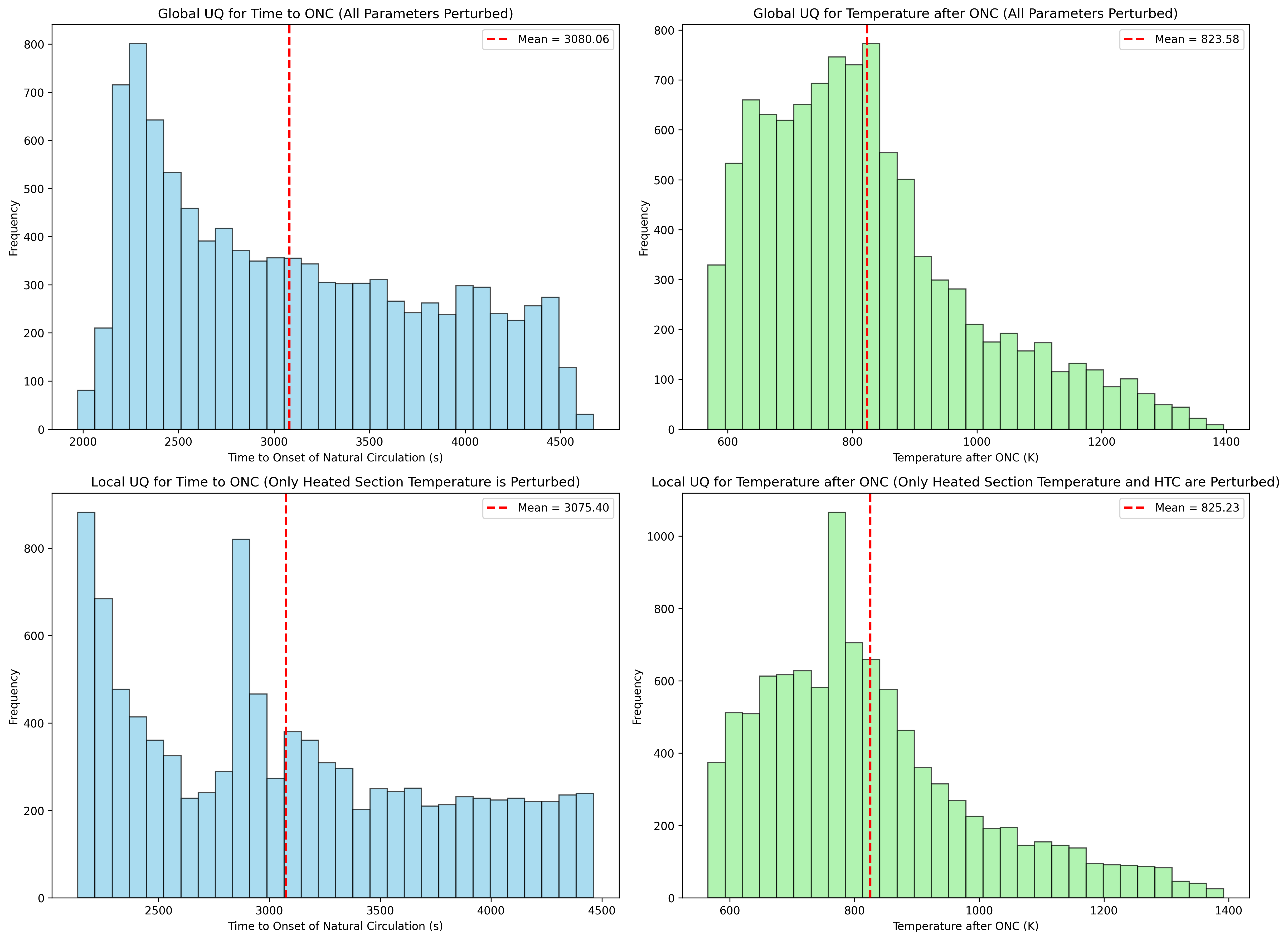}
\caption{Global and local uncertainty quantification reproduced from a previous work~\cite{eaheart2025sensitivity}}
\label{fig:uq}
\end{figure}

Following the sensitivity analysis, local and global uncertainty quantification were performed to assess the relative contribution of dominant versus non-dominant input parameters. For both outputs, 10,000 samples were propagated through a neural network surrogate under two conditions: all inputs varying and only the dominant inputs varying. The mean outputs were nearly identical across both conditions, confirming that the non-dominant parameters have minimal influence on the expected output, as shown in Figure~\ref{fig:uq}.

These findings motivate an evaluation of how multifidelity surrogate methods perform under different input configurations. In the remainder of this work, dominant inputs refer to the heated section temperature for time to ONC, and the heated section temperature and heat transfer coefficient for temperature after ONC. Non-dominant inputs refer to all remaining parameters for each respective output.

\subsubsection{Fidelity Level Definition}

\begin{figure}[h]
\centering
\includegraphics[width=0.85\textwidth]{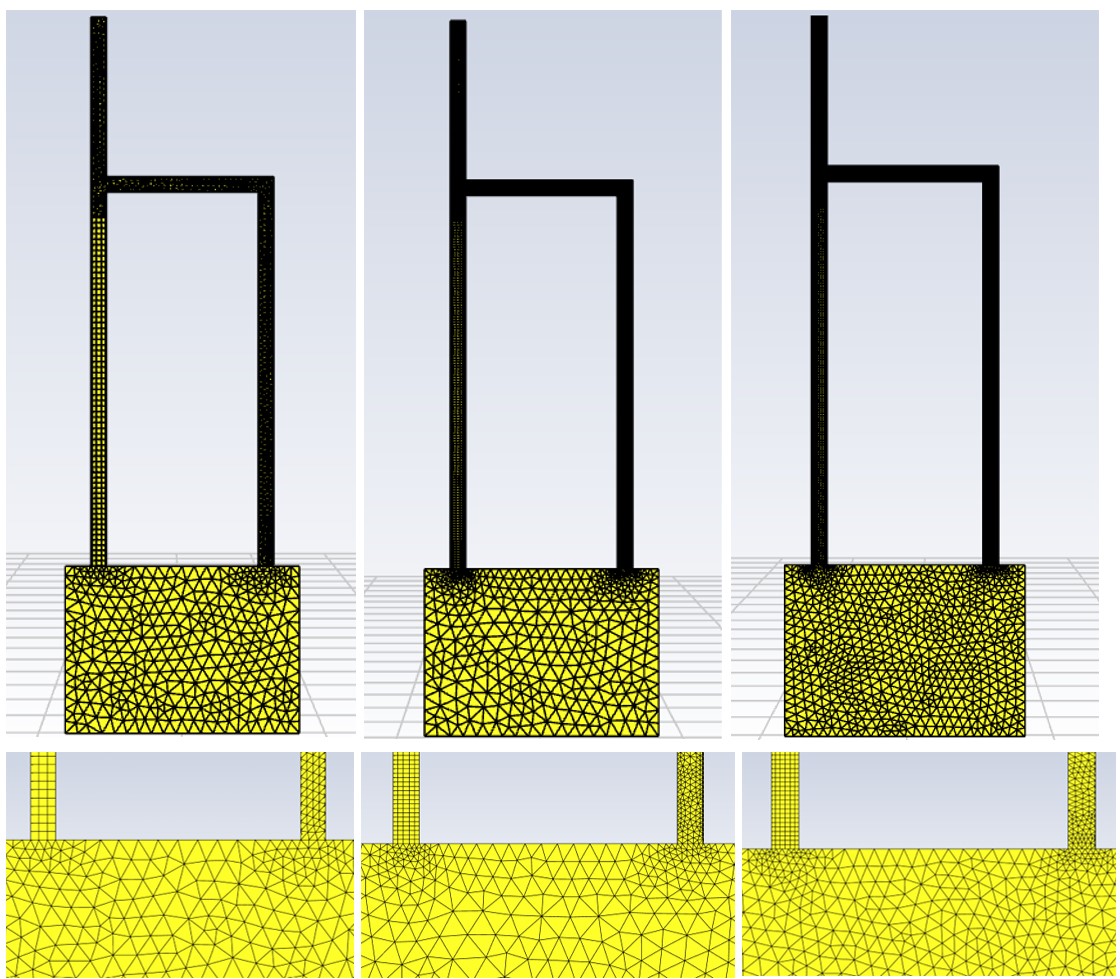}
\caption{Comparison of (left) low-fidelity (17.5k elements), (center) 
medium-fidelity (35k elements), and (right) high-fidelity (70k elements) 
mesh resolutions defining the fidelity hierarchy.}
\label{fig:meshes}
\end{figure}

To enable multifidelity machine learning studies, new simulation models were developed and generated at three mesh resolutions while keeping all solver settings, boundary conditions, physical models, and time stepping identical. The high-fidelity (HF) mesh contains approximately 70,000 elements, the medium-fidelity (MF) mesh contains approximately 35,000 elements, and the low-fidelity (LF) mesh contains approximately 17,500 elements. A total of 1000 simulations were generated for each fidelity level, which is double the size of the simulations that were available before \cite{eaheart2025sensitivity}. Figure~\ref{fig:meshes} shows the three mesh resolutions side-by-side. Each high-fidelity simulation sample requires approximately 6 days to complete the full transient when executed on a single processor. In comparison, the medium-fidelity and low-fidelity simulations require about 3 days and 1.5 days, respectively, underscoring the substantial computational cost associated with this problem.

\subsection{Multi-fidelity Machine Learning Methods}
\label{sec:mf-methods}

Several multifidelity machine learning methods were investigated in both two- and three-fidelity settings. These approaches are outlined below.

\paragraph{Multi-fidelity Neural Network: Delta Method}
The delta method models the high-fidelity response as a correction to a learned low-fidelity prediction. Following the generalized autoregressive multifidelity formulation in~\cite{meng2020composite}, the high-fidelity output is expressed as a function of both the input variables and the low-fidelity prediction, implemented here using a residual-based neural network approach. First, a neural network $f_L(x)$ is trained on the low-fidelity dataset. The trained model is evaluated at the high-fidelity input locations, and the discrepancy between the high-fidelity observations and these predictions is computed as
\begin{equation}
    r(x) = y_H(x) - f_L(x).
\end{equation}
A second neural network $f_{\Delta}(x, f_L(x))$ is then trained to model this residual, and the final prediction is
\begin{equation}
    \hat{y}_H(x) = f_L(x) + f_{\Delta}(x, f_L(x)).
\end{equation}
The two networks are trained sequentially, and this formulation does not require low- and high-fidelity datasets to share identical sampling locations.

\paragraph{Multi-fidelity Neural Network: Flag Method}
In the fidelity-flag approach (hereafter referred to as the \emph{flag method}), a single neural network is trained to learn multiple fidelity mappings simultaneously. This formulation belongs to the class of all-in-one architectures described in~\cite{guo2022multi}, in which different fidelity levels are modeled jointly within a unified network. The fidelity level is incorporated as an additional input feature by appending a fidelity indicator $f$ to the input vector. In the bi-fidelity case, $f \in \{0,1\}$, where $f=0$ denotes low-fidelity and $f=1$ denotes high-fidelity data. High-fidelity predictions are obtained by evaluating the learned mapping $F(x,f)$ at $f=1$,
\begin{equation}
    \hat{y}_H(x) = F(x,1).
\end{equation}
Correlations between fidelity levels are learned implicitly through shared network parameters. For $K$ fidelity levels, the indicator generalizes to $K$ categories (e.g., integer-valued or one-hot encoding).

\paragraph{Multi-fidelity Neural Network: Intermediate Method}
The Intermediate method~\cite{guo2022multi} is an all-in-one neural network in which the input variables are first processed through a shared trunk of hidden layers to produce a feature representation $h(x)$. From this representation, a low-fidelity prediction is generated,
\begin{equation}
    \hat{y}_L(x) = f_L\big(h(x)\big).
\end{equation}
This predicted low-fidelity value is then concatenated with the shared features to construct the high-fidelity prediction,
\begin{equation}
    \hat{y}_H(x) = f_H\big(h(x), \hat{y}_L(x)\big).
\end{equation}
Model parameters are optimized using the weighted loss
\begin{equation}
    \label{eq:weight_loss}
    \mathcal{L} = \alpha\,\mathrm{MSE}_{\mathrm{HF}} + (1-\alpha)\,\mathrm{MSE}_{\mathrm{LF}} + \lambda \lVert W \rVert_2^2,
\end{equation}
where $\alpha$ controls the relative weighting between fidelity levels and $\lambda$ is a regularization parameter. The term $\lambda \|W\|_2^2$ represents $L_2$ regularization (weight decay) and $W$ denotes the vector containing all trainable parameters of the neural network. $\mathrm{MSE}_{HF}$ is the mean squared error between the predicted and true high-fidelity outputs and $\mathrm{MSE}_{LF}$ represents the mean squared error for the low-fidelity predictions.

\paragraph{Multi-fidelity Neural Network: GPmimic Method}
The GPmimic method~\cite{guo2022multi} is an all-in-one architecture designed to emulate the linear correlation structure of autoregressive Gaussian process models while retaining the scalability of neural networks. The inputs are mapped through a shared neural network trunk to produce a latent representation $u(x)$, and a final linear mixing layer combines these latent features to generate predictions for each fidelity level,
\begin{equation}
    \hat{\mathbf{y}}(x) = W\,u(x) + b,
\end{equation}
where $W$ is a learned mixing matrix and $b$ is a bias term. Because no nonlinear activation is applied in the final layer, each fidelity output is a linear combination of the same shared features. As in the Intermediate method, the GPmimic method is trained using the weighted loss defined in Eq.\eqref{eq:weight_loss}.

\paragraph{Multi-fidelity Neural Network: Two-Step Method}
The two-step method~\cite{guo2022multi} trains separate neural networks sequentially. First, a neural network is trained on the low-fidelity dataset and evaluated at high-fidelity input locations to obtain low-fidelity predictions. A second neural network is then trained on the high-fidelity data to learn
\begin{equation}
    \hat{y}_H(x) = f_H\big(x, f_L(x)\big).
\end{equation}
Unlike the delta method, this approach directly learns the high-fidelity mapping as a function of the original inputs and the low-fidelity surrogate rather than explicitly modeling a residual correction.

\paragraph{Multi-fidelity Neural Network: Three-Step Method}
The three-step method~\cite{guo2022multi} extends the two-step formulation by separating linear and nonlinear inter-fidelity correlations. A neural network is first trained on the low-fidelity dataset. A linear network is then trained to capture dominant linear correlations between the predicted low-fidelity values and the high-fidelity response, after which a shallow nonlinear network models any remaining nonlinear relationships. The resulting prediction is
\begin{equation}
    \hat{y}_H(x) = f_H\big(x, f_L(x), y_{\mathrm{lin}}(x)\big).
\end{equation}

\paragraph{Multi-fidelity Gaussian Process (MF-GP)}
We also consider a multifidelity Gaussian process method in which the 
high-fidelity response is represented as a scaled low-fidelity function plus 
an independent discrepancy term~\cite{kennedy2000predicting},
\begin{equation}
    y_H(x) = \rho \, y_L(x) + \delta(x),
\end{equation}
where $\rho$ is a scalar scaling parameter and $\delta(x)$ is a Gaussian 
process capturing the residual between fidelity levels. The implementation 
follows a two-stage procedure: a Gaussian process is first fit to the 
low-fidelity data to obtain $\mu_L(x)$, after which $\rho$ is estimated via 
least-squares regression at the high-fidelity input locations. A second 
Gaussian process is then trained on the residuals $r(x) = y_H(x) - \rho \, 
\mu_L(x)$, yielding the final prediction $\hat{y}_H(x) = \rho \, \mu_L(x) + 
\mu_{\delta}(x)$.

\paragraph{Three-fidelity extensions}
Some of the multifidelity neural network methods considered in this work are inherently formulated for two fidelity levels and would require architectural modifications to incorporate additional fidelities. In contrast, the Intermediate, GPmimic, and Flag methods can be naturally extended to multiple fidelity levels without altering their overall network architecture, for example by adding additional outputs or intermediate representations. Accordingly, both two- and three-fidelity variants are evaluated for these methods, while the remaining methods are evaluated only in the bi-fidelity setting.

\subsection{Analytical Benchmarks}
\label{sec:analytical-benchmarks}

 \subsubsection{Two Fidelity Benchmarks}
 \label{sec:twofid-benchmarks}
Data to validate the two-fidelity methods' performance were generated using two-fidelity benchmark functions from the MF2 package~\cite{van2020mf2}. The benchmarks used, ordered by input dimensionality, were Forrester (1D), Booth (2D), Branin (2D), Park91A (4D), Hartmann (6D), and Borehole (8D). In the following definitions in this section, $f_h$ refers to the high-fidelity function, and $f_l$ refers to the low-fidelity function.

\paragraph{Bi-fidelity Forrester Function (1D)} 

The bi-fidelity Forrester function is derived from the classical one-dimensional Forrester test function~\cite{forrester2007multi} as shown below:
\begin{align}
f_h(x) &= (6x - 2)^2 \sin(12x - 4) \\
f_l(x) &= A f_h(x) + B(x - 0.5) + C
\end{align}
where $x \in [0,1]$. The recommended parameters are:
\[
A = 0.5, \quad B = 10, \quad C = -5
\]

\paragraph{Bi-fidelity Booth Function (2D)} 

The bi-fidelity Booth function is a two-input system of equations and is defined below~\cite{dong2015multi}:
\begin{align}
f_h(x_1, x_2) &= (x_1 + 2x_2 - 7)^2 + (2x_1 + x_2 - 5)^2 \\
f_l(x_1, x_2) &= f_h(0.4x_1, x_2) + 1.7x_1x_2 - x_1 + 2x_2
\end{align}
where
\[
x_1, x_2 \in [-10,10].
\]

\paragraph{Bi-fidelity Park91A Function (4D)} 
The bi-fidelity Park91A function is a four-dimensional benchmark function as shown below~\cite{xiong2013sequential}:
\begin{equation}
f_h(x_1,x_2,x_3,x_4)
=
\frac{x_1}{2}
\left(
\sqrt{1 + (x_2 + x_3^2)\frac{x_4}{x_1^2}} - 1
\right)
+
(x_1 + 3x_4)\exp\!\left(1 + \sin(x_3)\right)
\end{equation}

\begin{equation}
f_l(x_1,x_2,x_3,x_4)
=
\left(1 + \frac{\sin(x_1)}{10}\right) f_h(x_1,x_2,x_3,x_4)
- 2x_1 + x_2^2 + x_3^2 + 0.5
\end{equation}

\[
x_i \in [0,1], \quad i = 1,\dots,4
\]

\paragraph{Bi-fidelity Hartmann Function (6D)} 
The bi-fidelity Hartmann function is a six-dimensional set of equations~\cite{park2017remarks} defined by:
\begin{align}
f_h(x_1,\dots,x_6)
&=
-\frac{1}{1.94}
\left(
2.58
+
\sum_{i=1}^{4}
\alpha_i
\exp\!\left(
-\sum_{j=1}^{6} A_{ij}(x_j - P_{ij})^2
\right)
\right)
\\[6pt]
f_l(x_1,\dots,x_6)
&=
-\frac{1}{1.94}
\left(
2.58
+
\sum_{i=1}^{4}
\alpha'_i
f_{\exp}\!\left(
-\sum_{j=1}^{6} A_{ij}(x_j - P_{ij})^2
\right)
\right)
\end{align}

\begin{equation}
f_{\exp}(x)
=
\left(
\exp(-4/9)
+
\exp(-4/9)\,\frac{x+4}{9}
\right)^9
\end{equation}

These matrices and vectors are used in the Hartmann functions.
\[
A =
\begin{pmatrix}
10 & 3 & 17 & 3.5 & 1.7 & 8 \\
0.05 & 10 & 17 & 0.1 & 8 & 14 \\
3 & 3.5 & 1.7 & 10 & 17 & 8 \\
17 & 8 & 0.05 & 10 & 0.1 & 14
\end{pmatrix}
\]

\[
P = 10^{-4}
\begin{pmatrix}
1312 & 1696 & 5569 & 124 & 8283 & 5886 \\
2329 & 4135 & 8307 & 3736 & 1004 & 9991 \\
2348 & 1451 & 3522 & 2883 & 3047 & 6650 \\
4047 & 8828 & 8732 & 5743 & 1091 & 381
\end{pmatrix}
\]

\[
\alpha = \{1.0,\;1.2,\;3.0,\;3.2\},
\qquad
\alpha' = \{0.5,\;0.5,\;2.0,\;4.0\}
\]

\[
x_j \in [0,1], \quad j = 1,\dots,6
\]

\paragraph{Bi-fidelity Borehole Function (8D)}

The bi-fidelity Borehole benchmark ~\cite{xiong2013sequential} is constructed from the base function $f_b(x;A,B)$, defined as follows:
\[
f_b(x;\,A,B) = \frac{A \cdot T_u(H_u - H_l)}{\ln\!\left(\dfrac{r}{r_w}\right)
\left(B + \dfrac{2LT_u}{\ln\!\left(\dfrac{r}{r_w}\right) r_w^2 K_w} + 
\dfrac{T_u}{T_l}\right)}
\]
with $f_h(x) = f_b(x;\, A=2\pi,\, B=1)$, $f_l(x) = f_b(x;\, A=5,\, B=1.5)$, 
and input vector $x = (r_w, r, T_u, H_u, T_l, H_l, L, K_w)$. The input ranges are:
\[
r_w \in [0.05,0.15],\quad
r \in [100,50000],\quad
T_u \in [63070,115600],\quad
H_u \in [990,1110],
\]
\[
T_l \in [63.1,116],\quad
H_l \in [700,820],\quad
L \in [1120,1680],\quad
K_w \in [9855,12045].
\]

 \subsubsection{Three Fidelity Benchmarks}
 \label{sec:threefid-benchmarks}
 Data to validate the performance of the three-fidelity methods were generated using various three-fidelity benchmark functions~\cite{mainini2022analytical}. These benchmark functions include Forrester (1D), Rastrigin (2D and 5D), and Rosenbrock (2D and 5D).

\paragraph{Multi-fidelity Forrester (Adapted to Tri-Fidelity)} 
The multifidelity Forrester function is defined as shown below~\cite{mainini2022analytical,forrester2007multi}:
 \begin{align}
f_1(x) &= (6x - 2)^2 \sin(12x - 4), \\[6pt]
f_2(x) &= (5.5x - 2.5)^2 \sin(12x - 4), \\[6pt]
f_3(x) &= 0.75(6x - 2)^2 \sin(12x - 4) + 5(x - 0.5) - 2, \\[6pt]
f_4(x) &= 0.5(6x - 2)^2 \sin(12x - 4) + 10(x - 0.5) - 5,
\end{align}
where $x \in [0,1]$. Because only three-fidelities were necessary for this work, not all four defined fidelities were used in the creation of the benchmark functions. The ordering of the fidelities goes from highest ($f_1$) to lowest fidelity ($f_4$). For the purpose of this research, $f_1$ is not used, $f_2$ is treated as the high-fidelity function, $f_3$ is the medium-fidelity function, and $f_4$ is low-fidelity function.

\paragraph{Tri-fidelity Rosenbrock} 
The tri-fidelity Rosenbrock function is defined as shown below~\cite{mainini2022analytical,bryson2016variable,ficini2021assessing}:
\begin{equation}
f_1(\mathbf{x}) =
\sum_{i=1}^{D-1} \left[ 100\left(x_{i+1} - x_i^2\right)^2 + (1 - x_i)^2 \right]
\end{equation}

\begin{equation}
f_2(\mathbf{x}) = \sum_{i=1}^{D-1}\left[50\left(x_{i+1} - x_i^2\right)^2 + (-2 - x_i)^2\right] - \sum_{i=1}^{D} 0.5\,x_i
\end{equation}

\begin{equation}
f_3(\mathbf{x}) =
\frac{f_1(\mathbf{x}) - 4 - \sum_{i=1}^{D} 0.5\,x_i}
     {10 + \sum_{i=1}^{D} 0.25\,x_i}
\end{equation}

\[
\mathbf{x} \in [-2,2]^D
\]

Similarly to the representation in the Forrester function, $f_3$ represents the lowest fidelity, $f_2$ is the medium-fidelity, and $f_1$ is the high-fidelity solution. One interesting feature of these equations is that the number of input dimensions can be changed by refining the $D$ parameter. In this work, input dimensions of 2 and 5 were used to validate the performance of the three-fidelity methods.

\paragraph{Tri-fidelity Rastrigin}
The tri-fidelity shifted-rotated Rastrigin function~\cite{ficini2021assessing,wang2017generic} is defined as

\begin{equation}
f_i(\mathbf{z};\phi_i) = f_1(\mathbf{z}) + e_r(\mathbf{z}, \phi_i), \quad i=1,2,3,
\end{equation}
where the base Rastrigin function is
\begin{equation}
f_1(\mathbf{z}) = \sum_{j=1}^{D} \left( z_j^2 + 1 - \cos(10\pi z_j) \right),
\end{equation}
and
\begin{equation}
\mathbf{z} = R_D(\theta)(\mathbf{x} - \mathbf{x}^*).
\end{equation}

For the two-dimensional case,
\begin{equation}
R_2(\theta)=
\begin{pmatrix}
\cos\theta & -\sin\theta\\
\sin\theta & \cos\theta
\end{pmatrix},
\end{equation}
while for higher-dimensional cases the rotation operator can be extended using the Aguilera--Perez algorithm. The fidelity-dependent error term is
\begin{equation}
e_r(\mathbf{z}, \phi) =
\sum_{j=1}^{D} a(\phi)\cos^2\!\left(w(\phi) z_j + b(\phi) + \pi \right),
\end{equation}
with
\begin{align}
a(\phi) &= \Theta(\phi),\\
w(\phi) &= 10\pi\Theta(\phi),\\
b(\phi) &= 0.5\pi\Theta(\phi),\\
\Theta(\phi) &= 1 - 0.0001\phi.
\end{align}

The input domain is
\[
-0.1 \le x_i \le 0.2, \qquad i=1,\ldots,D,
\]
the rotation angle is $\theta=0.2$, and the optimum occurs at
\[
\mathbf{x}^* = (0.1,\ldots,0.1)^\top,
\]
for which the highest-fidelity objective value is zero. The fidelity parameters are
\[
\phi_1 = 10000 \quad \text{(high-fidelity)}, \qquad
\phi_2 = 5000 \quad \text{(medium-fidelity)}, \qquad
\phi_3 = 2500 \quad \text{(low-fidelity)}.
\]

\subsection{Onset of Natural Circulation (ONC) Experimental Design}
The objective of this research is to evaluate the performance of two- and three-fidelity machine learning methods introduced in Section~\ref{sec:mf-methods}. The evaluation is carried out using both analytical benchmark functions and the ONC CFD dataset consisting of 1000 low-, medium-, and high-fidelity simulations that we introduced in Section \ref{sec:data}.

To systematically evaluate how input sensitivity affects multifidelity performance, the experimental inputs are categorized into three sets: All Inputs, Dominant Inputs, and Non-Dominant Inputs. These definitions, summarized in Table~\ref{tab:input_definitions}, are derived from sensitivity analysis and uncertainty quantification~\cite{eaheart2025sensitivity} which identified the specific parameters driving the variance for each output quantity.

\begin{table}[h]
\centering
\caption{Definition of input parameter sets for the dominant and non-dominant sensitivity studies. The specific parameters included in the 'Dominant' and 'Non-Dominant' sets vary depending on the target output quantity, based on the sensitivity analysis results.}

\label{tab:input_definitions}
\begin{tabular}{p{3cm} p{5cm} p{6cm}}
\hline
\textbf{Target Output} & \textbf{Input Set} & \textbf{Included Parameters} \\
\hline
\multirow{3}{3cm}{Time to ONC} & All Inputs & All 8 input parameters \\
 & Dominant & Heated Section Temperature \\
 & Non-Dominant & All parameters \textit{excluding} Heated Section Temperature \\
\hline
\multirow{3}{3cm}{Temperature after ONC} & All Inputs & All 8 input parameters \\
 & Dominant & Heated Section Temperature and Unheated Section Heat Transfer Coefficient \\
 & Non-Dominant & All parameters \textit{excluding}  Heated Section Temperature and Unheated Section Heat Transfer Coefficient \\
\hline
\end{tabular}
\end{table}

The experimental workflow is designed to first validate each multifidelity method on the analytical benchmark functions described in Section~\ref{sec:analytical-benchmarks}. These benchmarks provide a controlled setting in which the behavior of the models can be assessed prior to application on the ONC dataset. Following benchmark validation, the methods are applied to the ONC data to evaluate their predictive performance under realistic conditions in an applied scenario.

\subsubsection{Benchmark Validation and ONC Application}

Each multifidelity method is first applied to the two- and three-fidelity benchmark functions to assess baseline performance. Hyperparameter tuning is performed using the benchmark problems, after which the same tuning procedure is repeated on the ONC dataset. The ONC hyperparameter tuning is conducted using all input parameters with time to ONC as a representative output variable. This setup helps prevent overfitting of each multifidelity method to specific sub-scenarios, enabling a fair comparison of all methods based on their generalization capability.

Once optimal hyperparameters are selected, the trained models are evaluated on multiple ONC prediction tasks. These include predicting time to ONC and temperature after ONC using different subsets of input variables: dominant inputs, non-dominant inputs, and the full input set. This approach enables a comparison of model sensitivity to input selection as well as an assessment of predictive robustness across fidelity configurations.

\subsubsection{Hyperparameter Tuning Strategy}
\label{sec:tuning_strategy}

Hyperparameter tuning is performed for each multifidelity method using a grid search approach. Tuning is first conducted on the analytical benchmark functions and subsequently applied to the ONC dataset. For neural network-based multifidelity methods, the shared hyperparameters and corresponding search ranges are summarized in Table~\ref{tab:hyperparameters}. The number of training epochs is kept fixed across all experiments to ensure convergence while limiting computational cost.

For the GPmimic and MFNN-Intermediate methods, additional method-specific hyperparameters are required. In the two-fidelity setting, a weighting parameter $\alpha$ is used to control the relative importance of the low- and high-fidelity data in the loss function, with $\alpha = 0.5$ corresponding to equal weighting of the two fidelities and the limiting cases $\alpha = 0$ and $\alpha = 1$ reducing to single-fidelity regression using only low- or high-fidelity data, respectively. A grid search is performed over $\alpha \in \{10^{-4}, 5\times10^{-4}, 10^{-3}, 5\times10^{-3}, 10^{-2}, 5\times10^{-2}\}$. In addition, a penalty parameter $\lambda$ is tuned over $\lambda \in \{10^{-1}, 10^{-5}, 10^{-4}, 5\times10^{-4}, 10^{-3}, 3\times10^{-3}\}$ following the formulation of the MFNN-Intermediate method.

In the three-fidelity setting, this weighting strategy is extended to account for low-, medium-, and high-fidelity data through a set of normalized weights $(w_l, w_m, w_h)$ satisfying $w_l + w_m + w_h = 1$. To reflect the hierarchical nature of the fidelities, a constrained grid search is performed with $w_h = 0.5, 0.6, 0.7$ and $w_m = 0.2, 0.3$, with $w_l$ determined implicitly. The penalty parameter $\lambda$ is tuned over a reduced grid given by $\lambda \in \{10^{-5}, 10^{-4}, 10^{-3}\}$. All hyperparameter configurations are selected based on the lowest RMSE, and the resulting optimal values for the benchmarks for each method are reported in Tables~\ref{tab:besthyperparameters} and~\ref{tab:fidelity_weights} and in Table~\ref{tab:best_config_real} for the ONC dataset.

For the MF-GP method, no explicit grid search is performed. Instead, the kernel structures are specified directly, using a Matern kernel with an additive white noise term for the low-fidelity process and an RBF kernel with white noise for the residual process. We explored various types of kernels before selecting these based on their strong performance. 

\begin{table}[h]
\centering
\caption{Hyperparameter tuning grid for Multi-fidelity Neural Networks.}
\label{tab:hyperparameters}
\begin{tabular}{ll}
\hline
\textbf{Hyperparameter} & \textbf{Search Values} \\ \hline
Number of Layers        & [2, 3, 4]              \\
Nodes per Layer         & [16, 32, 64, 128]      \\
Learning Rate           & [1e-4, 5e-4, 1e-3]     \\
Number of Epochs        & 500                    \\ \hline
\end{tabular}
\end{table}

\subsubsection{Two-fidelity vs. Three-fidelity Cost Matching Study}
\label{sec:Two-fidelity vs. Three-fidelity Cost Matching Study}

To evaluate the relative performance of two- and three-fidelity multifidelity methods, we designed a study focused on predicting high-fidelity simulation outputs. The primary objective of this comparison is to assess whether the inclusion of an intermediate fidelity level provides additional benefit beyond simply increasing the number of high-fidelity samples.

To ensure a fair comparison, computational cost is normalized across two-fidelity and three-fidelity experiments. Without cost matching, performance improvements could be attributed to increased computational expense rather than the use of additional fidelity levels. Because the high-fidelity simulations contain approximately four times as many mesh elements as the low-fidelity simulations, and roughly twice as many as the medium-fidelity simulations, relative costs are assigned as follows:
\begin{itemize}
    \item low-fidelity: 1
    \item medium-fidelity: 2  
    \item high-fidelity: 4
\end{itemize}

Using these normalized costs, total simulation budgets were defined and held constant across different fidelity configurations. In addition to comparing two-fidelity and three-fidelity setups at equal cost, all pairwise two-fidelity combinations (LF+HF, LF+MF, MF+HF) were also considered. This allowed us to assess whether performance improvements were due to the inclusion of an intermediate fidelity, specifically, or whether one fidelity level combination was inherently more informative than another. The computational budgets for these combinations can be seen in Table~\ref{tab:cost_matched_budgets}.

\begin{table}[htbp]
\centering
\caption{Cost-matched simulation budgets for two-fidelity (2F) and three-fidelity (3F) experiments. 
Costs are normalized relative to low-fidelity simulations (LF = 1, MF = 2, HF = 4).}
\label{tab:cost_matched_budgets}
\begin{tabular}{ccccc}
\hline
Fidelity Setup & $n_{\text{LF}}$ & $n_{\text{MF}}$ & $n_{\text{HF}}$ & Total Cost \\
\hline
\multicolumn{5}{c}{Total Cost = 300} \\
\hline
2F (LF + HF)  & 200  & 0    & 25  & 300 \\
2F (LF + MF)  & 200  & 50   & 0   & 300 \\
2F (MF + HF)  & 0    & 100  & 25  & 300 \\
3F (LF + MF + HF) & 150  & 50   & 12  & 298 \\
\hline
\multicolumn{5}{c}{Total Cost = 600} \\
\hline
2F (LF + HF)  & 400  & 0    & 50  & 600 \\
2F (LF + MF)  & 400  & 100  & 0   & 600 \\
2F (MF + HF)  & 0    & 200  & 50  & 600 \\
3F (LF + MF + HF) & 300  & 100  & 25  & 600 \\
\hline
\multicolumn{5}{c}{Total Cost = 1200} \\
\hline
2F (LF + HF)  & 800  & 0    & 100 & 1200 \\
2F (LF + MF)  & 800  & 200  & 0   & 1200 \\
2F (MF + HF)  & 0    & 400  & 100 & 1200 \\
3F (LF + MF + HF) & 600  & 200  & 50  & 1200 \\
\hline
\multicolumn{5}{c}{Total Cost = 1800} \\
\hline
2F (LF + HF)  & 1000 & 0    & 200 & 1800 \\
2F (LF + MF)  & 1000 & 400  & 0   & 1800 \\
2F (MF + HF)  & 0    & 500  & 200 & 1800 \\
3F (LF + MF + HF) & 1000 & 200  & 100 & 1800 \\
\hline
\end{tabular}
\end{table}

With these budgets in mind, separate dataset files were created for the high-fidelity train and test split to avoid data leakage. Because the maximum high-fidelity budget is 200 samples, the training set consists of 200 randomly selected samples from the full dataset, and the testing set consists of the remaining 800 samples. For each method, training is performed using the specified number of high-fidelity samples from this training subset, and evaluation is carried out on the fixed set of 800 high-fidelity test samples.

Similarly, for the LF + MF cases, separate dataset files are created for the medium-fidelity data to prevent data leakage. In this setting, the medium-fidelity data act as the highest fidelity level available. Since the maximum number of medium-fidelity training samples is 500, the dataset is split into 500 training samples and 500 testing samples. The methods are trained using the specified number of medium-fidelity samples from the training subset and are evaluated on the fixed set of 500 medium-fidelity test samples.

Although it may appear counterintuitive for the test set to be comparable in size to, or even larger than, the training set, this setup was intentionally adopted by the authors. It is important to note that this study includes a larger amount of high-fidelity data than is typically available in multi-fidelity research. Because the low-fidelity and high-fidelity datasets are of equal size (e.g., 1000 samples each), the objective is not to recover high-fidelity performance from a severely limited high-fidelity budget. Rather, the goal is to validate and compare the performance of the different methods against a reliable ground truth using a large reference dataset, thereby enabling a more robust assessment of which approaches may be most promising for similar problems. Accordingly, the availability of sufficient computational resources in this study allowed us to generate an equal number of high-fidelity samples, increasing confidence in the results and reporting of statistical metrics on a large sample pool.

\section{Results}
\label{sec:results}

\subsection{Benchmark Validation}
To validate the performance of the proposed two- and three-fidelity methods, all methods are first evaluated on a set of established multifidelity benchmark functions prior to application on engineering data. These benchmarks provide a controlled setting in which relative method behavior can be assessed independently of the complexities introduced by simulation data.

Hyperparameter tuning is performed independently for each method using the benchmark problems described in Section~\ref{sec:tuning_strategy}. A grid search is conducted over the shared neural network hyperparameters summarized in Table~\ref{tab:hyperparameters}. For each method, the root mean squared error (RMSE) is calculated across all benchmark functions, and the average RMSE is used as the selection criterion. The hyperparameter configuration yielding the lowest average RMSE is selected as the best-performing configuration for that method.

The resulting optimal hyperparameters for the two- and three-fidelity methods are summarized in Table~\ref{tab:besthyperparameters}. These configurations are fixed in subsequent benchmark evaluations to enable a fair comparison between methods.

\begin{table}[h]
\centering
\caption{Best hyperparameter configurations across two- and three-fidelity methods. All models were trained for 500 epochs. The RMSE is averaged across all benchmarks for each method.}
\label{tab:besthyperparameters}
\begin{tabular}{llccc}
\hline
Fidelity & Method & Hidden Layers & Learning Rate & Mean RMSE \\
\hline
2 & GPmimic & (128,128,128,128) & 0.001 & 38.5905 \\
2 & MF-NN-Delta & (64, 64, 64, 64) & 0.001 & \textbf{8.4437} \\
2 & MFNN-Intermediate & (128, 128, 128, 128) & 0.001 & 9.8412 \\
2 & MFNN-TwoStep & (64, 64, 64, 64) & 0.001 & 120.2910 \\
2 & MFNN-ThreeStep & (128,128,128,128) & 0.001 & 93.8388\\
2 & MFNN-Flag & (128,128,128,128) & 0.001 & 10.0238 \\
\hline
3 & GPmimic3f & (128,128,128,128) & 0.001 & 199.8821 \\
3 & MFNN-Intermediate3f & (128,128,128,128) & 0.001 & \textbf{98.7048} \\
3 & MFNN-Flag3f & (128,128,128,128) & 0.001 & 114.2706 \\
\hline
\end{tabular}
\end{table}

For the two- and three-fidelity GPmimic and intermediate methods, the optimal weighting and penalty parameters selected on the benchmarks are reported in Table~\ref{tab:fidelity_weights}. For these methods, the average RMSE obtained after tuning these parameters differs only slightly from that obtained in the first-stage hyperparameter selection. This difference reflects the use of different fidelity-weighting across the two stages. In the initial stage, a fixed value of $\alpha = 0.1$ was used to define the relative contribution of the fidelity levels. In the second stage, the tuning parameter ranges follow those of~\cite{guo2022multi}, which utilized smaller values of $\alpha$. With this additional tuning, the observed improvement in the average RMSE from the second stage was modest.

\begin{table}[h]
\centering
\caption{Optimal fidelity-weighting and penalty parameters selected on the benchmark problems for two- and three-fidelity methods. For three-fidelity models, weights satisfy $w_l + w_m + w_h = 1$.}
\label{tab:fidelity_weights}
\begin{tabular}{llcccccc}
\hline
Fidelity & Method & $\alpha$ & $w_h$ & $w_m$ & $w_l$ & $\lambda$ & Mean RMSE \\
\hline
2 & GPmimic & 0.05 & -- & -- & -- & 1e-05 & 38.4414 \\
2 & MFNN-Intermediate & 0.05 & -- & -- & -- & 0.1 & \textbf{8.3376} \\
\hline
3 & GPmimic3f & -- & 0.5 & 0.2 & 0.3 & 0.0001 & 160.2200 \\
3 & MFNN-Intermediate3f & -- & 0.7 & 0.2 & 0.1 & 0.001 & \textbf{75.4586} \\
\hline
\end{tabular}
\end{table}

The best-performing hyperparameter configurations identified during tuning were then applied to each method and trained for 2000 epochs. Among the two-fidelity approaches, the two-step and three-step methods exhibit consistently weaker performance and are therefore omitted for brevity. The benchmark RMSE results for the remaining two-fidelity methods are summarized in Table~\ref{tab:rmse_comparison}.

\begin{table}[h]
\centering
\caption{RMSE comparison across benchmarks for two-fidelity methods, ordered by input dimension.}
\label{tab:rmse_comparison}
\begin{tabular}{lccccc}
\hline
Benchmark & GPmimic & MF-GP & MF-NN-Delta & MFNN-Flag & MFNN-Interm. \\
\hline
Forrester (1D)  & 0.0411 & \textbf{0.0000} & 0.0370 & 0.1320 & 0.0827 \\
Booth (2D)  & 32.7409 & \textbf{0.0067} & 13.1901 & 25.3527 & 25.9151 \\
Branin (2D)  & 14.7299 & \textbf{0.0154} & 8.9195 & 8.3862 & 9.9333 \\
Park91A (4D)  & 0.1610 & \textbf{0.0116} & 0.1629 & 0.2407 & 0.1358 \\
Hartmann6 (6D)  & 0.1529 & \textbf{0.1028} & 0.1125 & 0.1512 & 0.1550 \\
Borehole (8D)  & 3.2645 & \textbf{0.0820} & 5.3271 & 4.6495 & 4.9388 \\
\hline
\end{tabular}
\end{table}

As shown in Table~\ref{tab:rmse_comparison}, among the two-fidelity methods, the MF-GP method exhibits the strongest performance, achieving the lowest RMSE across all benchmark functions. Among the neural network-based approaches, no single method consistently outperforms the others. The MFNN-Delta method achieves the lowest error on the Forrester, Booth, and Hartmann6 benchmarks, while MFNN-Flag performs best on the Branin and Borehole problems, and MFNN-Intermediate provides the lowest RMSE on the Park91A benchmark.

To qualitatively assess predictive behavior beyond aggregate error metrics, parity and residual plots are shown for two representative benchmark functions. 

\begin{figure}[h]
\centering
\includegraphics[width=\linewidth]{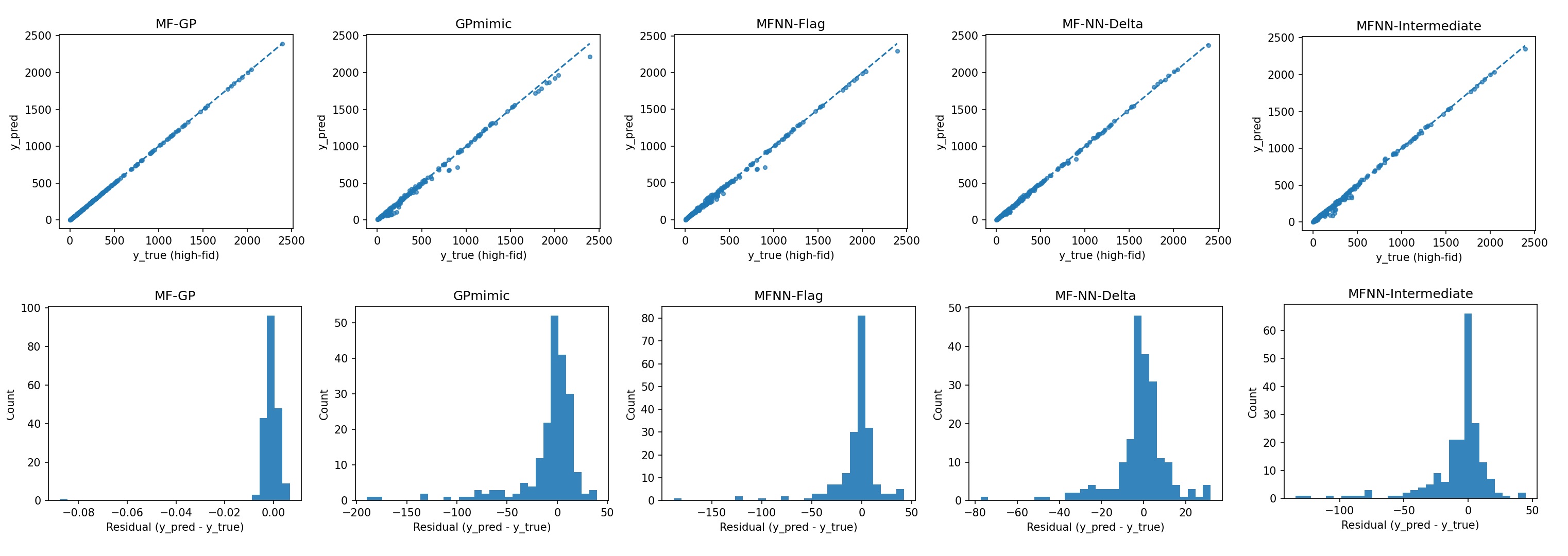}
\caption{Parity plot comparing predicted and true high-fidelity outputs (top) and residual distribution (bottom) for the Booth (2D) benchmark. Labels: y\_pred = predicted value, y\_true=ground truth.}
\label{fig:booth_parity_residuals}
\end{figure}

The Booth function is shown as a two-dimensional benchmark, with parity plots and residual distributions presented in Figure~\ref{fig:booth_parity_residuals}. Due to the simplicity of the underlying input-output relationship, all methods exhibit strong agreement between predicted and true high-fidelity values, with points closely located along the parity line. The residual distributions are narrowly centered around zero, indicating low variance across methods. Subtle differences are visible in the spread of residuals, with MF-GP exhibiting the tightest distribution, consistent with its substantially lower RMSE value reported in Table~\ref{tab:rmse_comparison}.

\begin{figure}[h]
\centering
\includegraphics[width=\linewidth]{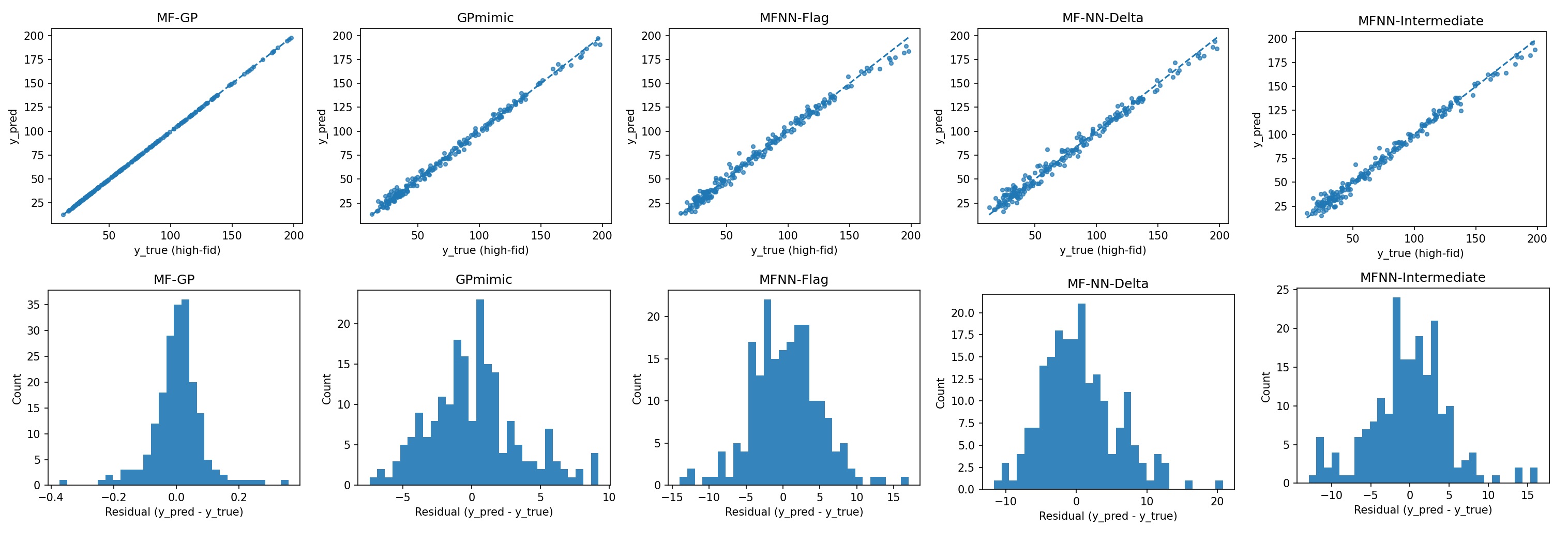}
\caption{Parity plot comparing predicted and true high-fidelity outputs (top) and residual distribution (bottom) for the Borehole (8D) benchmark. Labels: y\_pred = predicted value, y\_true=ground truth}
\label{fig:borehole_parity_residuals}
\end{figure}

Figure~\ref{fig:borehole_parity_residuals} shows the corresponding parity and residual plots for the Borehole benchmark. The Borehole function is chosen to show method behavior in a higher-dimensional setting with more complex input–output interactions. As shown in Figure~\ref{fig:borehole_parity_residuals}, increased scatter is observed in the parity plots relative to the Booth case. The residual distributions are correspondingly broader, indicating higher variance in prediction errors across all methods. Despite this increased complexity, MF-GP maintains relatively well-centered residuals and more uniform error distributions compared to the other neural network-based approaches, suggesting more stable performance as dimensionality increases.

The RMSE results for the three-fidelity benchmarks are summarized in Table~\ref{tab:rmse_comparison_3f}. Overall, the MFNN-Intermediate method demonstrates the strongest predictive performance, achieving the lowest RMSE on four of the five benchmark functions. Specifically, it provides the best results for the Forrester (1D), Rastrigin (2D), Rastrigin (5D), and Rosenbrock (2D) problems. The GPmimic method remains competitive and yields the lowest error only for the Rosenbrock (5D) benchmark. In contrast, the MFNN-Flag method consistently produces higher prediction errors across all evaluated functions. As expected, prediction difficulty increases with dimensionality and problem complexity, with the Rosenbrock (5D) case representing the most challenging benchmark for all methods.

\begin{table}[h]
\centering
\caption{RMSE comparison across benchmarks for three-fidelity methods.}
\label{tab:rmse_comparison_3f}
\begin{tabular}{lccc}
\hline
Benchmark & GPmimic & MFNN-Intermediate & MFNN-Flag \\
\hline
Forrester 1D & 0.0417 & \textbf{0.0340} & 0.0606 \\
Rastrigin 2D & 0.1509 & \textbf{0.0718} & 0.0855 \\
Rastrigin 5D & 0.5669 & \textbf{0.5302} & 0.7285 \\
Rosenbrock 2D & 10.6742 & \textbf{9.6140} & 24.7987 \\
Rosenbrock 5D & \textbf{281.1450} & 317.0519 & 410.2674 \\
\hline
\end{tabular}
\end{table}

To qualitatively assess the predictive behavior of these methods across varying degrees of complexity, parity and residual plots are presented for two representative benchmarks: Rosenbrock (2D) and Rastrigin (5D). 



\begin{figure}[h]
\centering
\includegraphics[width=0.7\linewidth]{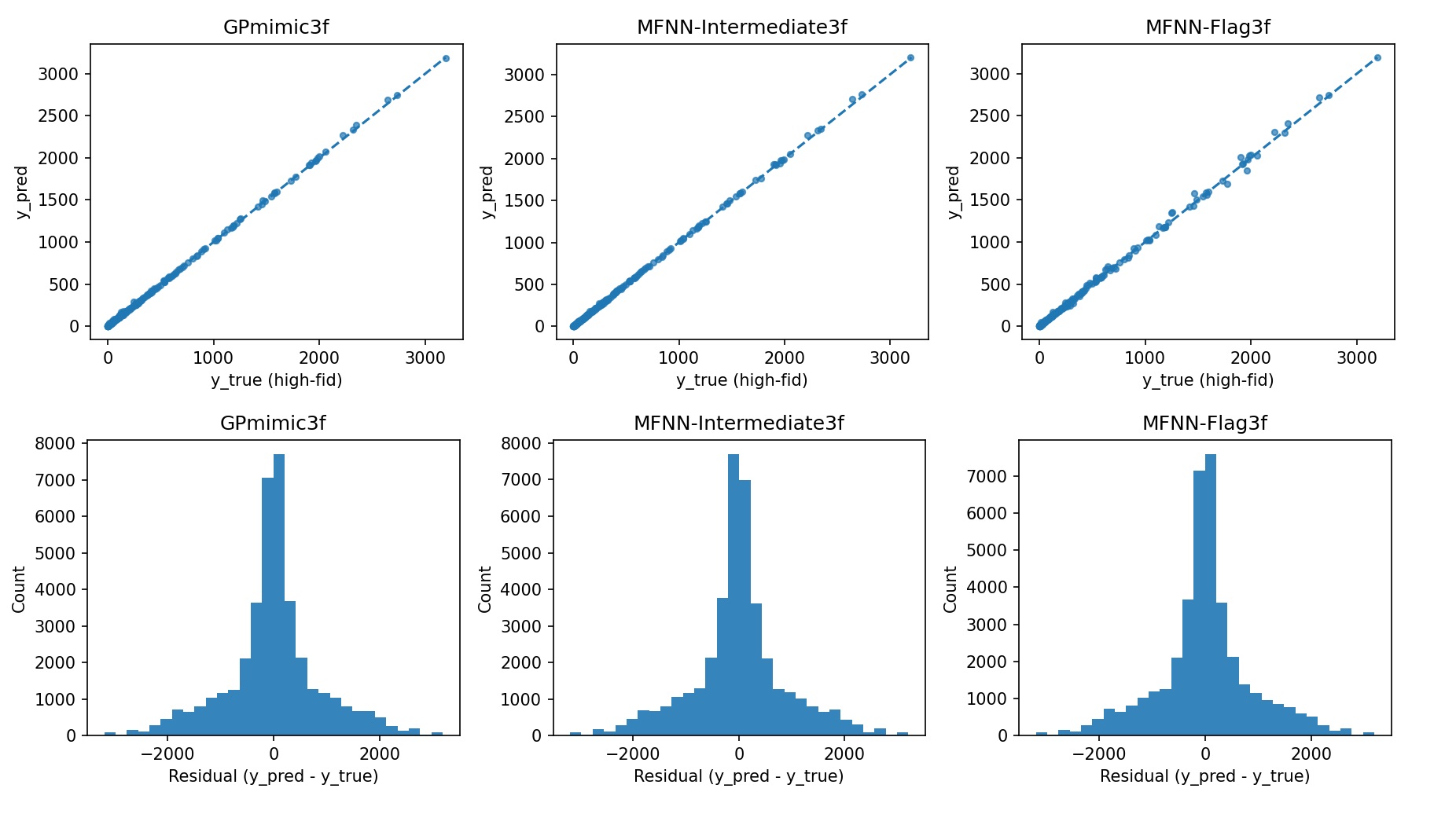}
\caption{Parity plot comparing predicted and true high-fidelity outputs (top) and residual distribution (bottom) for the Rosenbrock (2D) benchmark on three-fidelity methods. Labels: y\_pred = predicted value, y\_true=ground truth}
\label{fig:rosenbrock_2d}
\end{figure}

The results for the two-dimensional Rosenbrock function are presented in Figure~\ref{fig:rosenbrock_2d}. The parity plots in the top row of Figure~\ref{fig:rosenbrock_2d} show that the GPmimic3f and MFNN-Intermediate3f methods perform similarly, with predicted values adhering closely to the parity line. In contrast, the MFNN-Flag3f method exhibits slightly worse performance; the points are not as tight to the diagonal and show visible scatter. This trend is confirmed by the residual distributions in the bottom row of Figure~\ref{fig:rosenbrock_2d}, where the Flag3f residuals are less tightly clustered around zero compared to the sharp peaks observed for the other two methods.

\begin{figure}[h]
\centering
\includegraphics[width=0.7\linewidth]{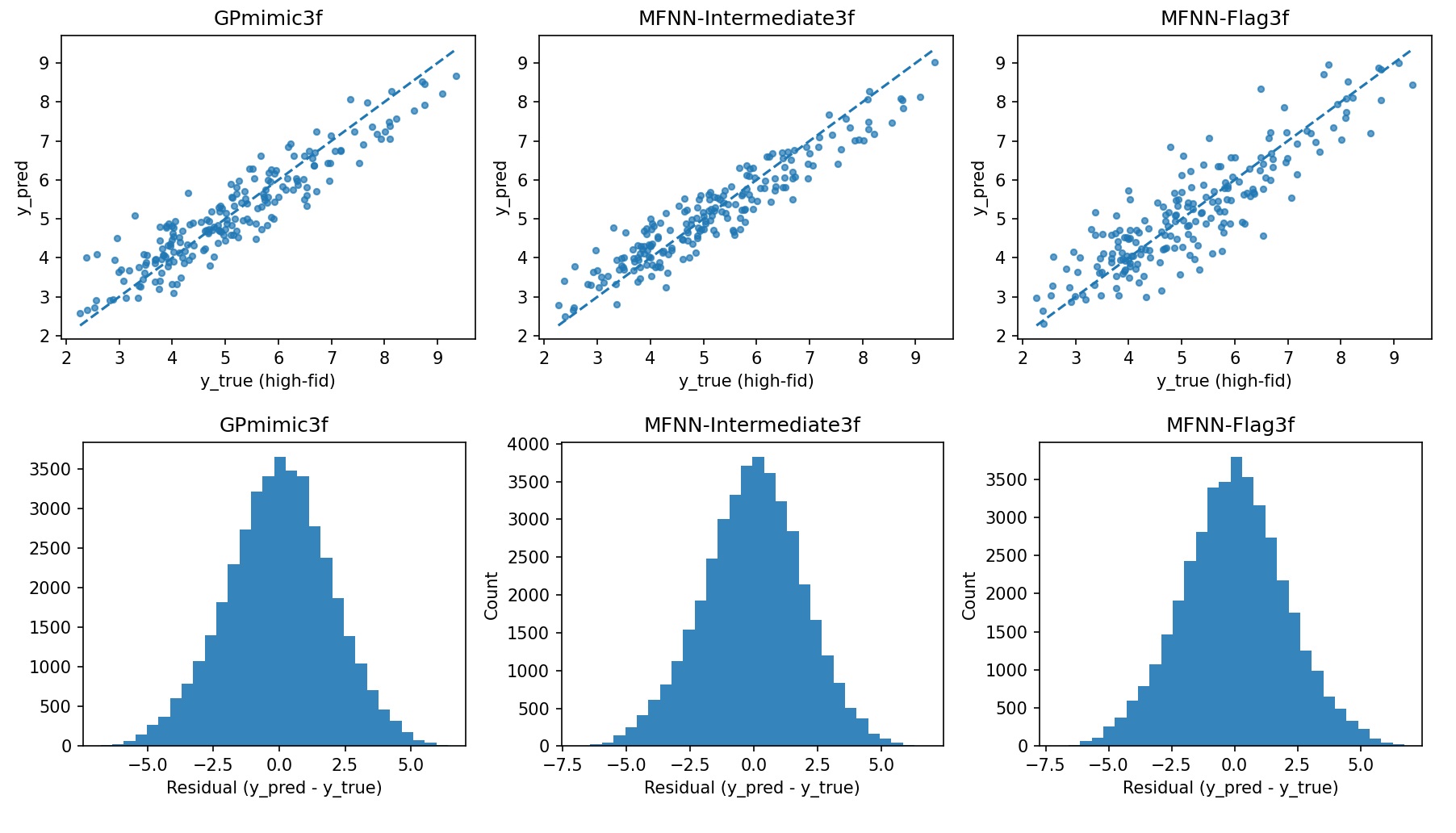}
\caption{Parity plot comparing predicted and true high-fidelity outputs (top) and residual distribution (bottom) for the Rastrigin (5D) benchmark on three-fidelity methods.}
\label{fig:rastrigin_5d}
\end{figure}

Finally, the results for the five-dimensional Rastrigin function are presented in Figure~\ref{fig:rastrigin_5d}. The parity plots in the top row show that the predictions are much more scattered compared to the tight lines seen in the Rosenbrock 2D results. The points do not fall as perfectly on the diagonal line. The residual histograms (bottom row) confirm this, showing a wider spread of errors rather than the sharp peaks clustered at zero seen in the previous benchmarks. Visually, all three methods exhibit similar qualitative behavior, although MFNN-Flag shows somewhat larger errors.

\subsection{Simulation Data Hyperparameter Tuning}

After the methods were validated on the analytical benchmarks, hyperparameter tuning was performed on the ONC data generated in this research. The same hyperparameter tuning grid used for the benchmark functions (Table~\ref{tab:hyperparameters}) was applied to the ONC dataset. Hyperparameter tuning was performed on all of the input parameters and time to ONC as the output. The resulting hyperparameter configurations can be seen in Table~\ref{tab:best_config_real}.

\begin{table}[h]
\centering
\caption{Best hyperparameter configurations on the real dataset (All $\rightarrow$ Time to ONC) across two- and three-fidelity methods. All methods were trained for 500 epochs.}
\label{tab:best_config_real}
\begin{tabular}{llcccc}
\hline
Fidelity & Method & Layers & Nodes per Layer & Learning Rate & RMSE (seconds) \\
\hline
2 & GPmimic & 3 & 128 & 0.0010 & 386.8805 \\
2 & MF-NN-Delta & 4 & 128 & 0.0010 & \textbf{194.1512} \\
2 & MFNN-Flag & 4 & 128 & 0.0010 & 194.7106 \\
2 & MFNN-Intermediate & 4 & 128 & 0.0010 & 199.8040 \\
2 & MFNN-ThreeStep & 2 & 128 & 0.0010 & 2977.1743 \\
2 & MFNN-TwoStep & 3 & 32 & 0.0010 & 3065.7572 \\
\hline
3 & GPmimic3f & 4 & 128 & 0.0010 & 382.8426 \\
3 & MFNN-Flag3f & 3 & 128 & 0.0010 & \textbf{288.1738} \\
3 & MFNN-Intermediate3f & 3 & 64 & 0.0010 & 378.6047 \\
\hline
\end{tabular}
\end{table}

Comparing these results with the benchmark configurations in Table~\ref{tab:besthyperparameters}, several differences are observed. For the two-fidelity methods, MFNN-Delta widened from 64 to 128 nodes per layer, while GPmimic reduced depth from four to three layers. The Two-Step and Three-Step methods showed the most pronounced reductions in architecture size. For the three-fidelity methods, GPmimic3f retained its benchmark architecture, while MFNN-Intermediate3f reduced to three layers of 64 nodes and MFNN-Flag3f reduced to three layers of 128 nodes.

The relative performance trends also differ between datasets. For the two-fidelity case, MFNN-Delta achieved the lowest RMSE on the benchmark data and also yields the lowest RMSE on the ONC dataset, although MFNN-Flag and MFNN-Intermediate perform comparably. In contrast to the benchmark results, where MFNN-Delta exhibited clearer dominance among the neural network methods, the ONC dataset shows more similar performance among MFNN-Delta, MFNN-Flag, and MFNN-Intermediate. As in the benchmark experiments, the Two-Step and Three-Step methods remain the poorest performers by a substantial margin. GPmimic continues to outperform the Two-Step and Three-Step methods but performs worse than the stronger multifidelity neural network variants.

For the three-fidelity methods, the benchmark ranking was MFNN-Intermediate3f, followed by MFNN-Flag3f, and then GPmimic3f. On the ONC dataset, this ordering changes: MFNN-Flag3f achieves the lowest RMSE, followed by MFNN-Intermediate3f, while GPmimic3f again performs worst among the three methods.

To ensure consistent comparisons across varying experimental conditions, the optimal hyperparameter configurations identified in Table~\ref{tab:best_config_real} were held fixed for all subsequent analyses, including the dominant and non-dominant input studies. This ensures that observed performance differences are attributable to fidelity combinations and input subsets rather than hyperparameter variation. Consistent with their substantially higher error across both datasets, the MFNN-TwoStep and MFNN-ThreeStep methods are omitted from the remainder of this study. 

The second-stage tuning of fidelity-weighting and penalty parameters ($\alpha$ and $\lambda$) was not repeated for the ONC dataset. Given that this stage produced only modest improvements on the benchmark problems, the parameter values selected during benchmark tuning were carried forward and applied directly to the ONC analyses. No additional hyperparameter tuning was performed for the temperature after onset of natural circulation output parameter. The same hyperparameter configurations obtained from the time-to-ONC tuning case were used for this output and for all combinations of input parameters.

\subsection{ONC Results at Fixed Computational Budget}

Table~\ref{tab:lfhf_1800} shows the performance for the LF+HF configuration at a fixed computational budget of 1800 as defined in Table~\ref{tab:cost_matched_budgets}. For the full input set, the two-fidelity MF-GP method achieves the highest predictive accuracy for both outputs. For \textit{Time to ONC}, MF-GP (2F) attains an RMSE of 102.35 with $R^2 = 0.9796$, exceeding the performance of all other methods. For \textit{Temperature after ONC}, it achieves near-perfect performance (RMSE = 1.08, $R^2 = 1.0$), again outperforming all alternatives. However, this accuracy comes with substantially longer training times. For all inputs, MF-GP requires 392.29 seconds for \textit{Time to ONC} and 227.39 seconds for \textit{Temperature after ONC}, whereas the MFNN-based approaches generally require much less time. Thus, while MF-GP provides the strongest predictive performance, it is also the most computationally intensive method in this configuration.

\begin{table}[t]
\centering
\caption{Performance comparison for the LF+HF configuration at a fixed total computational budget of 1800. The Time (s) column refers to the computatiuonal time needed by each method}
\label{tab:lfhf_1800}

\begin{tabular}{lccccccc}
\hline
 &  & \multicolumn{3}{c}{\textbf{Time to ONC (Seconds)}} & \multicolumn{3}{c}{\textbf{Temp. after ONC (K)}} \\
\cline{3-5} \cline{6-8}
\textbf{Inputs} & \textbf{Method} 
& RMSE & $R^2$ & Time (s) 
& RMSE & $R^2$ & Time (s) \\
\hline

\multicolumn{8}{l}{\textbf{All Inputs}} \\
\hline
 & GPmimic (2F) & 312.59 & 0.8100 & 5.57 & 110.96 & 0.6103 & 5.75 \\
 & GPmimic3f (3F) & 286.92 & 0.8399 & 86.94 & 95.88 & 0.7090 & 73.92 \\
 & MF-GP (2F) & 102.35 & 0.9796 & 392.29 & 1.08 & 1.0000 & 227.39 \\
 & MF-NN-Delta (2F) & 139.13 & 0.9624 & 8.60 & 25.14 & 0.9800 & 8.94 \\
 & MFNN-Flag (2F) & 152.88 & 0.9546 & 5.48 & 44.22 & 0.9381 & 5.62 \\
 & MFNN-Flag3f (3F) & 206.43 & 0.9171 & 44.53 & 47.61 & 0.9283 & 38.51 \\
 & MFNN-Intermediate (2F) & 193.52 & 0.9272 & 9.81 & 23.09 & 0.9831 & 8.44 \\
 & MFNN-Intermediate3f (3F) & 323.65 & 0.7963 & 59.07 & 67.74 & 0.8548 & 49.77 \\
\hline
\multicolumn{8}{l}{\textbf{Dominant Inputs}} \\
\hline
 & GPmimic (2F) & 142.57 & 0.9605 & 6.44 & 96.44 & 0.7056 & 5.72 \\
 & GPmimic3f (3F) & 119.82 & 0.9721 & 123.93 & 17.63 & 0.9902 & 72.28 \\
 & MF-GP (2F) & 140.97 & 0.9614 & 155.89 & 1.11 & 1.0000 & 84.70 \\
 & MF-NN-Delta (2F) & 116.37 & 0.9737 & 9.83 & 2.70 & 0.9998 & 8.63 \\
 & MFNN-Flag (2F) & 115.69 & 0.9740 & 5.52 & 20.91 & 0.9862 & 5.66 \\
 & MFNN-Flag3f (3F) & 130.10 & 0.9671 & 19.41 & 34.15 & 0.9631 & 39.04 \\
 & MFNN-Intermediate (2F) & 116.41 & 0.9737 & 10.09 & 4.70 & 0.9993 & 10.18 \\
 & MFNN-Intermediate3f (3F) & 115.06 & 0.9743 & 37.61 & 11.96 & 0.9955 & 51.16 \\
\hline
\multicolumn{8}{l}{\textbf{NonDominant Inputs}} \\
\hline
 & GPmimic (2F) & 773.75 & -0.1640 & 5.57 & 191.48 & -0.1604 & 5.68 \\
 & GPmimic3f (3F) & 866.96 & -0.4614 & 114.52 & 194.88 & -0.2019 & 96.92 \\
 & MF-GP (2F) & 314.07 & 0.8082 & 306.49 & 102.34 & 0.6686 & 277.37 \\
 & MF-NN-Delta (2F) & 1005.18 & -0.9645 & 9.21 & 233.09 & -0.7195 & 8.38 \\
 & MFNN-Flag (2F) & 756.93 & -0.1140 & 5.66 & 187.05 & -0.1073 & 5.76 \\
 & MFNN-Flag3f (3F) & 770.75 & -0.1550 & 49.40 & 187.27 & -0.1099 & 47.24 \\
 & MFNN-Intermediate (2F) & 400.63 & 0.6879 & 9.51 & 148.62 & 0.3009 & 9.80 \\
 & MFNN-Intermediate3f (3F) & 917.75 & -0.6376 & 60.30 & 200.77 & -0.2757 & 72.45 \\
\hline

\end{tabular}
\end{table}


When restricted to dominant inputs, several two-fidelity methods achieve similarly strong performance for \textit{Time to ONC}. In this configuration, MFNN-Flag (2F) attains the lowest RMSE (115.69) with $R^2 = 0.9740$, while MFNN-Delta (2F) and MFNN-Intermediate (2F) achieve nearly identical performance with RMSE values of 116.37 and 116.41 and $R^2 = 0.9737$ for both. MF-GP (2F) performs slightly worse with RMSE = 140.97 ($R^2 = 0.9614$), while GPmimic (2F) yields RMSE = 142.57 ($R^2 = 0.9605$). Notably, the neural network-based approaches achieve these results with runtimes of approximately 5–10 seconds, compared to 155.89 seconds for MF-GP.

For \textit{Temperature after ONC}, MF-GP (2F) again achieves the highest accuracy ($R^2 = 1.0000$, RMSE = 1.11). However, MFNN-Delta (2F) and MFNN-Intermediate (2F) also achieve extremely high accuracy with RMSE values of 2.70 and 4.70 and $R^2$ values of 0.9998 and 0.9993, respectively, while requiring substantially lower runtimes (approximately 8–10 seconds). MFNN-Flag (2F) also performs well, achieving RMSE = 20.91 with $R^2 = 0.9862$ and a runtime of approximately 5.7 seconds. These results indicate that for dominant inputs, several two-fidelity neural network approaches achieve accuracy comparable to MF-GP while providing significantly lower computational cost.

Performance decreases significantly when only non-dominant inputs are used. For \textit{Time to ONC}, nearly all methods produce near-zero or negative $R^2$ values, with the exception of MF-GP (2F), which maintains fair predictive capability ($R^2 = 0.8082$). MFNN-Intermediate (2F) also achieves moderately positive performance ($R^2 = 0.6879$), while all other approaches yield negligible or negative $R^2$ values. A similar pattern is observed for \textit{Temperature after ONC}, where MF-GP (2F) achieves $R^2 = 0.6686$, followed by MFNN-Intermediate (2F) at $R^2 = 0.3009$, while all other methods remain near or below zero. Although MF-GP provides the best predictive capability for the non-dominant inputs, it again requires substantially longer training times (306.49 seconds for \textit{Time to ONC} and 277.37 seconds for \textit{Temperature after ONC}). These results are not surprising, as the non-dominant inputs play a negligible role in predicting the target outputs per the sensitvity analysis study. This case therefore serves as a sanity check, confirming that all methods fail to accurately predict the outputs under such conditions. \textit{Accordingly, the non-dominant input cases will not be discussed further.}

Within the LF+HF configuration, two-fidelity implementations generally provide competitive or superior accuracy compared to their corresponding three-fidelity variants at the same computational budget. For example, under all inputs for \textit{Time to ONC}, MFNN-Intermediate (2F) achieves $R^2 = 0.9272$ compared to 0.7963 for MFNN-Intermediate3f (3F). However, GPmimic shows a slight improvement when incorporating the third fidelity level, achieving $R^2 = 0.8399$ compared to 0.8100 for the two-fidelity variant. Across the input groupings and both output variables, three-fidelity variants generally incur substantially longer runtimes while providing only modest accuracy improvements or, in some cases, reduced predictive performance relative to the corresponding two-fidelity models.

Tables~\ref{tab:lfhf_1800}, \ref{tab:lfmf_1800}, and \ref{tab:mfhf_1800} compare predictive performance across fidelity pairings at a fixed computational budget of 1800. Among the three configurations, the LF+HF pairing provides the strongest overall performance for \textit{Time to ONC}. With all or dominant inputs available, several methods achieve very high accuracy, with $R^2$ values exceeding 0.95 and RMSE values substantially lower than those observed for the other fidelity combinations. In contrast, the LF+MF configuration exhibits considerably weaker predictive capability for \textit{Time to ONC}, with $R^2$ values generally below 0.50 even when all inputs are used and declining to near-zero or negative values when only non-dominant inputs are available. The MF+HF pairing demonstrates intermediate behavior, achieving strong performance for all and dominant inputs.

For \textit{Temperature after ONC}, the behavior is notably different. Across all fidelity pairings, MF-GP consistently achieves near-perfect predictive accuracy with $R^2$ values approaching 1.0 for both all-input and dominant-input configurations. Neural network approaches such as MFNN-Delta, MFNN-Flag, and MFNN-Intermediate also achieve very high accuracy while requiring substantially shorter runtimes. Overall, these results indicate that the predictive accuracy of the models depends strongly on both the fidelity pairing and the availability of informative input variables, with the LF+HF configuration providing the most reliable performance for the more challenging \textit{Time to ONC} prediction task.

\subsection{Effect of Increasing Budget}

Table~\ref{tab:lfhf_scaling_dominant} shows the scaling behavior of the LF+HF (2F) configuration for dominant inputs as the total computational budget increases from 300 to 1800. Across methods, predictive performance for \textit{Time to ONC} remains consistently strong, with several approaches achieving $R^2$ values above 0.97 even at the smallest budget. MF-GP achieves $R^2 = 0.9739$ at a budget of 300 and maintains similar accuracy across larger budgets. Neural network approaches including MFNN-Delta, MFNN-Flag, and MFNN-Intermediate also achieve high and stable performance across budgets, with $R^2$ values consistently near or above 0.97. For \textit{Temperature after ONC}, MF-GP again achieves the highest accuracy across all budgets with $R^2$ values approaching 1.0, while the neural network methods also provide highly accurate predictions with substantially shorter runtimes.

\setlength{\tabcolsep}{4pt}
\renewcommand{\arraystretch}{1.1}
\small
\begin{table}[t]
\centering
\caption{Scaling behavior for LF+HF (2F) configuration across increasing total budgets for \textbf{dominant inputs}.}
\label{tab:lfhf_scaling_dominant}

\begin{tabular}{lccccccc}
\hline
 &  & \multicolumn{3}{c}{\textbf{Time to ONC}} & \multicolumn{3}{c}{\textbf{Temp. After ONC}} \\
\cline{3-5} \cline{6-8}
\textbf{Budget} & \textbf{Method}
& RMSE & $R^2$ & Time (s)
& RMSE & $R^2$ & Time (s) \\
\hline

\multicolumn{8}{l}{\textbf{Total Budget = 300}} \\
\hline
 & GPmimic & 149.23 & 0.9567 & 22.42 & 95.30 & 0.7125 & 4.10 \\
 & MF-GP & 115.79 & 0.9739 & 7.61 & 6.24 & 0.9988 & 2.60 \\
 & MF-NN-Delta & 118.81 & 0.9726 & 40.56 & 21.47 & 0.9854 & 6.44 \\
 & MFNN-Flag & 129.59 & 0.9673 & 23.81 & 38.42 & 0.9533 & 3.68 \\
 & MFNN-Intermediate & 120.38 & 0.9718 & 43.01 & 9.60 & 0.9971 & 7.10 \\
\hline
\multicolumn{8}{l}{\textbf{Total Budget = 600}} \\
\hline
 & GPmimic & 122.17 & 0.9710 & 37.88 & 99.88 & 0.6843 & 4.43 \\
 & MF-GP & 119.15 & 0.9724 & 19.87 & 2.05 & 0.9999 & 7.44 \\
 & MF-NN-Delta & 113.39 & 0.9750 & 72.76 & 4.99 & 0.9992 & 7.14 \\
 & MFNN-Flag & 116.70 & 0.9735 & 35.22 & 35.84 & 0.9593 & 4.17 \\
 & MFNN-Intermediate & 114.73 & 0.9744 & 71.59 & 8.40 & 0.9978 & 7.86 \\
\hline
\multicolumn{8}{l}{\textbf{Total Budget = 1200}} \\
\hline
 & GPmimic & 151.09 & 0.9556 & 35.17 & 97.60 & 0.6985 & 5.09 \\
 & MF-GP & 128.29 & 0.9680 & 90.09 & 1.76 & 0.9999 & 52.05 \\
 & MF-NN-Delta & 112.09 & 0.9756 & 50.80 & 3.00 & 0.9997 & 7.90 \\
 & MFNN-Flag & 116.74 & 0.9735 & 39.04 & 18.80 & 0.9888 & 4.99 \\
 & MFNN-Intermediate & 114.02 & 0.9747 & 75.01 & 4.18 & 0.9994 & 8.69 \\
\hline
\multicolumn{8}{l}{\textbf{Total Budget = 1800}} \\
\hline
 & GPmimic & 142.57 & 0.9605 & 6.44 & 96.44 & 0.7056 & 5.72 \\
 & MF-GP & 140.97 & 0.9614 & 155.89 & 1.11 & 1.0000 & 84.70 \\
 & MF-NN-Delta & 116.37 & 0.9737 & 9.83 & 2.70 & 0.9998 & 8.63 \\
 & MFNN-Flag & 115.69 & 0.9740 & 5.52 & 20.91 & 0.9862 & 5.66 \\
 & MFNN-Intermediate & 116.41 & 0.9737 & 10.09 & 4.70 & 0.9993 & 10.18 \\
\hline
\end{tabular}
\end{table}

MFNN-Delta, MFNN-Flag, and MFNN-Intermediate achieve consistently high $R^2$ values for \textit{Time to ONC} across all budgets, typically exceeding 0.97. GPmimic generally performs worse than the other methods but still maintains positive $R^2$ values across budgets. Overall, the results indicate that when informative input variables are available, several multifidelity neural network approaches achieve performance comparable to MF-GP while requiring substantially lower training times.

Table~\ref{tab:lfhf_scaling_all} shows the corresponding scaling behavior when all inputs are used. In this case, increasing the computational budget generally leads to improved predictive accuracy across several methods. This observation is logical as more samples are typically needed to inform a larger feature space, while for 1-2 dominant inputs, the lowest computational budget was sufficient.

\setlength{\tabcolsep}{4pt}
\renewcommand{\arraystretch}{1.1}
\small
\begin{table}[t]
\centering
\caption{Scaling behavior for LF+HF (2F) configuration across increasing total budgets for \textbf{all inputs}.}
\label{tab:lfhf_scaling_all}

\begin{tabular}{lccccccc}
\hline
 &  & \multicolumn{3}{c}{\textbf{Time to ONC}} & \multicolumn{3}{c}{\textbf{Temp After ONC}} \\
\cline{3-5} \cline{6-8}
\textbf{Budget} & \textbf{Method}
& RMSE & $R^2$ & Time (s)
& RMSE & $R^2$ & Time (s) \\
\hline

\multicolumn{8}{l}{\textbf{Total Budget = 300}} \\
\hline
 & GPmimic & 670.01 & 0.1272 & 3.90 & 200.77 & -0.2757 & 3.91 \\
 & MF-GP & 137.84 & 0.9631 & 6.15 & 8.95 & 0.9975 & 5.61 \\
 & MF-NN-Delta & 301.93 & 0.8228 & 6.43 & 80.15 & 0.7967 & 6.54 \\
 & MFNN-Flag & 177.22 & 0.9389 & 3.68 & 86.25 & 0.7645 & 3.44 \\
 & MFNN-Intermediate & 295.27 & 0.8305 & 6.56 & 75.48 & 0.8197 & 6.77 \\
\hline
\multicolumn{8}{l}{\textbf{Total Budget = 600}} \\
\hline
 & GPmimic & 583.04 & 0.3391 & 3.21 & 126.20 & 0.4959 & 4.52 \\
 & MF-GP & 121.94 & 0.9711 & 64.25 & 3.89 & 0.9995 & 49.70 \\
 & MF-NN-Delta & 225.78 & 0.9009 & 5.07 & 48.82 & 0.9246 & 7.00 \\
 & MFNN-Flag & 163.07 & 0.9483 & 3.22 & 50.12 & 0.9205 & 4.06 \\
 & MFNN-Intermediate & 237.29 & 0.8905 & 5.76 & 69.54 & 0.8470 & 7.66 \\
\hline
\multicolumn{8}{l}{\textbf{Total Budget = 1200}} \\
\hline
 & GPmimic & 315.24 & 0.8068 & 3.88 & 117.01 & 0.5667 & 5.73 \\
 & MF-GP & 126.16 & 0.9691 & 262.79 & 1.72 & 0.9999 & 141.07 \\
 & MF-NN-Delta & 149.51 & 0.9565 & 5.89 & 33.29 & 0.9649 & 9.60 \\
 & MFNN-Flag & 151.25 & 0.9555 & 3.59 & 52.18 & 0.9138 & 6.07 \\
 & MFNN-Intermediate & 217.59 & 0.9079 & 6.47 & 26.35 & 0.9780 & 10.18 \\
\hline
\multicolumn{8}{l}{\textbf{Total Budget = 1800}} \\
\hline
 & GPmimic & 312.59 & 0.8100 & 5.57 & 110.96 & 0.6103 & 5.75 \\
 & MF-GP & 102.35 & 0.9796 & 392.29 & 1.08 & 1.0000 & 227.39 \\
 & MF-NN-Delta & 139.13 & 0.9624 & 8.60 & 25.14 & 0.9800 & 8.94 \\
 & MFNN-Flag & 152.88 & 0.9546 & 5.48 & 44.22 & 0.9381 & 5.62 \\
 & MFNN-Intermediate & 193.52 & 0.9272 & 9.81 & 23.09 & 0.9831 & 8.44 \\
\hline
\end{tabular}
\end{table}

For \textit{Time to ONC} in Table~\ref{tab:lfhf_scaling_all}, MF-GP generally achieves the strongest performance across budgets, with $R^2$ increasing from 0.9631 at a budget of 300 to 0.9796 at 1800. Neural network methods also show improvements with increasing budget, particularly MFNN-Delta and MFNN-Flag, which achieve $R^2$ values above 0.95 at higher budgets. For \textit{Temperature after ONC}, MF-GP again achieves near-perfect accuracy across all budgets, while the neural network approaches also achieve strong predictive performance with substantially lower runtimes.

\begin{figure}[t]
    \centering
    \includegraphics[width=0.4\linewidth]{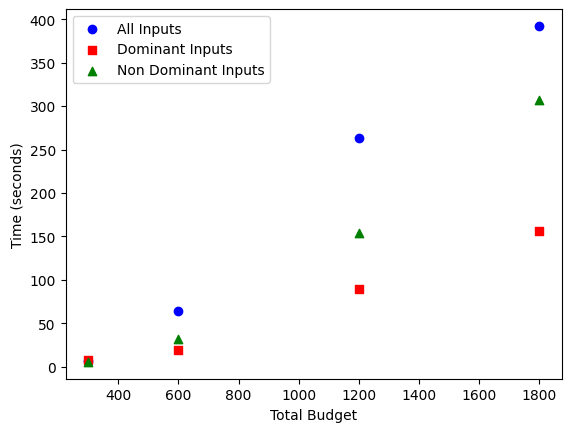}
    \caption{MF-GP runtime as a function of total computational budget for different input selections and Time to ONC output. Results are shown for all inputs, dominant inputs, and non-dominant inputs.}
    \label{fig:MFGP_costs}
\end{figure}

Figure~\ref{fig:MFGP_costs} further illustrates the scaling behavior of MF-GP runtime as a function of total computational budget for different input selections. As the computational budget increases, MF-GP training time grows substantially. For a fixed budget, runtime is highest when all inputs are included, lower when non-dominant inputs are used, and lowest when only dominant inputs are included. Thus, both increasing computational budget and increasing input dimensionality lead to higher MF-GP training times. In contrast, the neural network methods maintain substantially lower runtimes across budgets. These results suggest that when dominant input variables are known, strong multifidelity predictive performance can be achieved even at relatively small computational budgets, reducing the need for large training budgets.

\section{Discussions}
\label{sec:discussion}

Our analysis and results demonstrate that model performance strongly depends on both the dataset and the inputs used. Methods perform differently across input subsets (all, dominant, and non-dominant) and across output quantities, and strong performance on benchmark problems does not necessarily translate to engineering data. For example, several neural network approaches that perform less competitively  in benchmarks performed well for the ONC application. In contrast, MF-GP maintains comparatively strong performance across input groupings and in both cases for the benchmarks and the engineering problem. This variation demonstrates that the effectiveness of a modeling approach depends not only on the method itself but also on the strength of the underlying input–output relationships in the data.

Across experiments, the dominant-input configuration consistently provided the strongest predictive performance, frequently outperforming models trained on the full set of inputs. This finding suggests that including additional non-informative variables can introduce noise that weakens the learned relationships, while also increasing the amount of data required for training, which is particularly costly when high-fidelity simulations are involved. In contrast, models trained only on non-dominant inputs frequently struggled to capture the underlying relationships, highlighting the importance of input informativeness in multifidelity modeling.

These findings highlight the importance of evaluating multiple modeling strategies rather than relying on performance in a single benchmark configuration. Strong results on one dataset or input subset do not guarantee similar behavior under different input selections or data characteristics. In this study, models performed well when informative inputs were available but struggled when the input–output relationship was weak or when only non-dominant inputs were used.

The comparison of fidelity configurations further underscores that performance is problem-dependent. In this study, two-fidelity LF+HF configurations consistently outperformed their three-fidelity counterparts at a fixed computational budget. However, this does not imply that three-fidelity approaches are inherently inferior. Rather, the results suggest that the contribution of each fidelity level depends on the information it provides relative to the others. The LF+HF pairing produced stronger performance than either LF+MF or MF+HF, particularly for \textit{Time to ONC}. This pattern indicates that the medium-fidelity data, as constructed in this research, did not yield the same predictive gains as the low- and high-fidelity combination.

One possible explanation is that the medium-fidelity representation may provide limited additional information beyond what is already captured by the low- and high-fidelity levels. These findings should not be interpreted as a general limitation of three-fidelity approaches but as a feature of these specific data. With different fidelity definitions, stronger cross-level correlations, or alternative allocation of computational budget, three-fidelity methods may yield different outcomes. These results show that the effectiveness of multifidelity modeling depends not only on the number of fidelity levels but also on the informativeness and interaction between those levels.

Scaling analysis reveals a clear trade-off between predictive accuracy and computational cost. MF-GP achieves strong and consistent performance across challenging non-dominant input subsets, particularly as budget increases, but its runtime grows with both computational budget and input dimensionality. Neural network approaches, while generally less robust with limited information, offer significantly lower training times. The scaling experiments also show that when dominant inputs are known, accurate predictions can often be achieved even with relatively small computational budgets.

Lastly, the authors observed that most of the multifidelity methods evaluated in this study achieved reliable performance in at least one experimental setup, with the exception of MFNN-TwoStep and MFNN-ThreeStep, which demonstrated significantly weaker performance. In addition, the GPmimic method showed generally consistent performance across most runs; however, some cases exhibited moderate variability depending on the random seed. These observations suggest that the method may occasionally converge to suboptimal model parameters, highlighting a degree of sensitivity to stochastic initialization. Although most runs yielded comparable results, the presence of such outliers indicates that GPmimic may exhibit limited stability, especially at lower training budgets.

\section{Conclusions}
\label{sec:conclusions}
This work investigated the use of multifidelity machine learning surrogate models to predict the time to onset of natural circulation (ONC) and the temperature after ONC for a high-temperature gas-cooled reactor (HTGR) depressurized loss of forced cooling transient. High-fidelity computational fluid dynamics simulations (CFD) were generated using an Ansys Fluent model, and two additional fidelity levels were created by systematically coarsening the computational mesh to one-half and one-quarter of the high-fidelity resolution. Several multifidelity surrogate modeling methods were evaluated, including Gaussian processes and multiple neural network-based approaches, first validated on analytical benchmark functions before application to the ONC CFD dataset.

To assess how input sensitivity affects multifidelity performance, the study leveraged prior sensitivity analysis results to partition the input parameters into dominant, non-dominant, and full input sets. Across experiments, the dominant-input configuration consistently produced the strongest predictive performance, frequently outperforming models trained using the full set of inputs. This result suggests that including weakly informative variables can introduce noise that degrades surrogate model accuracy, whereas focusing on the most influential parameters strengthens the learned input–output relationships. A cost-matched comparison was also conducted to evaluate the performance of different fidelity pairings under equal computational budgets. The results showed that the LF+HF configuration generally produced stronger performance than either the LF+MF or MF+HF combinations. In addition, three-fidelity models did not consistently outperform their two-fidelity counterparts at equal computational cost, indicating that the inclusion of additional fidelity levels does not automatically improve predictive performance.

Overall, the results demonstrate that the effectiveness of multifidelity surrogate models depends strongly on input informativeness, the relationship between fidelity levels, and the allocation of computational resources. In this study, MF-GP provided the most robust performance when limited information was available, while neural network-based approaches offered competitive accuracy with substantially lower training times when informative inputs were known. Future work could extend this framework by incorporating spatiotemporal data to enable surrogate modeling of transient field quantities rather than scalar outputs. Such approaches may allow prediction of system evolution without requiring full high-fidelity transient simulations, which remain computationally expensive for large-scale reactor safety analyses. Additionally, more interpretable machine learning approaches, such as Kolmogorov–Arnold Networks (KANs) \cite{panczyk2025opening}, can be employed to derive symbolic expressions that improve transparency and interpretability in multifidelity modeling.

\section*{Data Availability}
\label{sec:data_avail}

The authors have all the data and codes to reproduce all the results in this work currently in a private GitHub repository. To ensure the confidentiality of this research, the authors will make this repository public during an advanced stage of the review process, which will be listed under our research group's public Github page: \url{https://github.com/aims-umich}

\section*{Acknowledgment}
This research was sponsored by the Department of Energy Office of Nuclear Energy's University Nuclear Leadership Program (UNLP) Graduate Fellowship.

\section*{CRediT Author Statement}

\begin{itemize}

    \item \textbf{Meredith Eaheart}: Conceptualization, Methodology, Software, Validation, Formal Analysis, Visualization, Investigation, Data Curation, Funding Acquisition, Writing - Original Draft.
    \item \textbf{Majdi I. Radaideh}: Conceptualization, Methodology, Software, Validation, Data Curation, Funding Acquisition, Supervision, Project Administration, Writing - Review and Edit. 

\end{itemize}

\bibliographystyle{elsarticle-num}
\setlength{\bibsep}{0pt plus 0.3ex}
{\small
\bibliography{master_reference}}

\clearpage
\appendix
\section{Appendix}

Tables \ref{tab:lfmf_1800}–\ref{tab:mfhf_1800} complement the results presented in Table \ref{tab:lfhf_1800} by reporting the LF+MF (2F) and MF+HF (2F) prediction metrics, respectively. For brevity, rows corresponding to non-dominant inputs are omitted, as these models were previously observed to exhibit weak performance in Table \ref{tab:lfhf_1800}, as expected from the sensitivity analysis study.

\begin{table}[!ht]
\centering
\caption{Performance comparison for the LF+MF (2F) configuration at a total computational budget of 1800.}
\label{tab:lfmf_1800}

\begin{tabular}{lccccccc}
\hline
 &  & \multicolumn{3}{c}{\textbf{Time to ONC}} & \multicolumn{3}{c}{\textbf{Temp after ONC}} \\
\cline{3-5} \cline{6-8}
\textbf{Inputs} & \textbf{Method}
& RMSE & $R^2$ & Time (s)
& RMSE & $R^2$ & Time (s) \\
\hline

\multicolumn{8}{l}{\textbf{All Inputs}} \\
\hline
 & GPmimic (2F) & 555.05 & 0.3888 & 4.52 & 103.57 & 0.6490 & 5.90 \\
 & MF-GP (2F) & 527.92 & 0.4471 & 495.54 & 0.89 & 1.0000 & 295.25 \\
 & MF-NN-Delta (2F) & 684.92 & 0.0693 & 6.76 & 14.09 & 0.9935 & 9.34 \\
 & MFNN-Flag (2F) & 513.92 & 0.4760 & 4.68 & 36.75 & 0.9558 & 5.90 \\
 & MFNN-Intermediate (2F) & 529.50 & 0.4438 & 8.83 & 13.05 & 0.9944 & 10.54 \\
\hline
\multicolumn{8}{l}{\textbf{Dominant Inputs}} \\
\hline
 & GPmimic (2F) & 528.55 & 0.4458 & 6.07 & 68.63 & 0.8459 & 6.35 \\
 & MF-GP (2F) & 747.30 & -0.1079 & 52.56 & 0.93 & 1.0000 & 100.21 \\
 & MF-NN-Delta (2F) & 554.06 & 0.3910 & 9.12 & 3.02 & 0.9997 & 9.57 \\
 & MFNN-Flag (2F) & 526.86 & 0.4493 & 6.00 & 13.40 & 0.9941 & 6.14 \\
 & MFNN-Intermediate (2F) & 535.92 & 0.4302 & 10.95 & 3.39 & 0.9996 & 11.21 \\
\hline

\end{tabular}
\end{table}

\begin{table}[!ht]
\centering
\caption{Performance comparison for the MF+HF (2F) configuration at a total computational budget of 1800.}
\label{tab:mfhf_1800}

\begin{tabular}{lccccccc}
\hline
 &  & \multicolumn{3}{c}{\textbf{Time to ONC}} & \multicolumn{3}{c}{\textbf{Temp after ONC}} \\
\cline{3-5} \cline{6-8}
\textbf{Inputs} & \textbf{Method}
& RMSE & $R^2$ & Time (s)
& RMSE & $R^2$ & Time (s) \\
\hline

\multicolumn{8}{l}{\textbf{All Inputs}} \\
\hline
 & GPmimic (2F) & 401.26 & 0.6869 & 5.03 & 84.21 & 0.7756 & 5.00 \\
 & MF-GP (2F) & 416.79 & 0.6622 & 80.26 & 0.97 & 1.0000 & 69.38 \\
 & MF-NN-Delta (2F) & 135.73 & 0.9642 & 8.01 & 23.39 & 0.9827 & 7.40 \\
 & MFNN-Flag (2F) & 189.38 & 0.9303 & 4.52 & 39.91 & 0.9496 & 4.79 \\
 & MFNN-Intermediate (2F) & 334.65 & 0.7823 & 8.82 & 30.84 & 0.9699 & 8.88 \\
\hline
\multicolumn{8}{l}{\textbf{Dominant Inputs}} \\
\hline
 & GPmimic (2F) & 156.54 & 0.9524 & 3.95 & 178.19 & -0.0049 & 5.05 \\
 & MF-GP (2F) & 467.46 & 0.5751 & 11.21 & 1.03 & 1.0000 & 14.50 \\
 & MF-NN-Delta (2F) & 115.92 & 0.9739 & 7.25 & 2.84 & 0.9997 & 7.70 \\
 & MFNN-Flag (2F) & 112.59 & 0.9754 & 4.71 & 16.40 & 0.9915 & 4.86 \\
 & MFNN-Intermediate (2F) & 119.74 & 0.9721 & 8.82 & 3.15 & 0.9997 & 8.85 \\
\hline

\end{tabular}
\end{table}

\end{document}